\documentclass[sigconf,nonacm]{acmart} 

\usepackage{multirow}
\usepackage{soul}
\usepackage{etoolbox}
\usepackage{mathtools}
\usepackage{enumitem}

\usepackage{algorithmic}
\usepackage[ruled,vlined,noend]{algorithm2e}

\usepackage[noabbrev,capitalise]{cleveref}

\usepackage[binary-units,per-mode=symbol]{siunitx}

\usepackage{subcaption}

\AtBeginDocument{%
  \providecommand\BibTeX{{%
    \normalfont B\kern-0.5em{\scshape i\kern-0.25em b}\kern-0.8em\TeX}}}

\copyrightyear{2021} 
\acmYear{2021}
\setcopyright{acmcopyright}\acmConference[GECCO '21]{2021 Genetic and Evolutionary Computation Conference}{July 10--14, 2021}{Lille, France}
\acmBooktitle{2021 Genetic and Evolutionary Computation Conference (GECCO '21), July 10--14, 2021, Lille, France}
\acmPrice{15.00}
\acmDOI{10.1145/3449639.3459273}
\acmISBN{978-1-4503-8350-9/21/07}

\begin{document}

\title[A Signal-Centric Perspective on the Evolution of Symbolic Communication]{A Signal-Centric Perspective on the Evolution of Symbolic Communication}{}



\author{Quintino Francesco Lotito}
\affiliation{%
  \institution{University of Trento}
  \city{Trento}
  \country{Italy}
}
\orcid{0000-0001-7084-8339}
\email{quintino.lotito@studenti.unitn.it}

\author{Leonardo Lucio Custode}
\affiliation{%
  \institution{University of Trento}
  \city{Trento}
  \country{Italy}
}
\orcid{0000-0002-1652-1690}
\email{leonardo.custode@unitn.it}

\author{Giovanni Iacca}
\affiliation{%
  \institution{University of Trento}
  \city{Trento}
  \country{Italy}
}
\orcid{0000-0001-9723-1830}
\email{giovanni.iacca@unitn.it}

\renewcommand{\shortauthors}{Lotito et al.}


\begin{abstract}
The evolution of symbolic communication is a longstanding open research question in biology. While some theories suggest that it originated from sub-symbolic communication (i.e., iconic or indexical), little experimental evidence exists on how organisms can actually evolve to define a shared set of symbols with unique interpretable meaning, thus being capable of encoding and decoding discrete information. Here, we use a simple synthetic model composed of sender and receiver agents controlled by Continuous-Time Recurrent Neural Networks, which are optimized by means of neuro-evolution. We characterize signal decoding as either regression or classification, with limited and unlimited signal amplitude. First, we show how this choice affects the complexity of the evolutionary search, and leads to different levels of generalization. We then assess the effect of noise, and test the evolved signaling system in a referential game. In various settings, we observe agents evolving to share a dictionary of symbols, with each symbol spontaneously associated to a 1-D unique signal. Finally, we analyze the constellation of signals associated to the evolved signaling systems and note that in most cases these resemble a Pulse Amplitude Modulation system.
\end{abstract}

\ccsdesc[500]{Computing methodologies~Artificial life}
\ccsdesc[500]{Computing methodologies~Multi-agent systems}
\ccsdesc[300]{Computing methodologies~Cooperation and coordination}
\ccsdesc[300]{Computing methodologies~Neural networks}
\ccsdesc[300]{Theory of computation~Multi-agent learning}


\keywords{Symbolic Communication, Neuro-evolution, NEAT, Continuous-Time Recurrent Neural Networks, Agent-Based Model.}

\maketitle


\section{Introduction}\label{sec:intro}

\ifdefined\IEEEPARstart{\IEEEPARstart{T}{}he}\else{The}\fi{ } origin of language is considered as the last known major transition in evolution \cite{smith1997major}. Language is in turn perceived as the culmination of an evolutionary history that led from the iconic/indexical forms of sub-symbolic communication to a symbolic one \cite{deacon1998symbolic,plotkin2001major}. Unravelling this history by gathering biological evidence is, however, extremely hard. Yet, studying the evolution of symbolic communication is a fundamental step toward understanding how language originated. One possible way to gain insights into this evolutionary process consists in analyzing how primordial forms of symbolic communication can evolve \textit{in silico}, in agent-based models. Previous literature has shown that, under given conditions, it is possible to evolve various forms of symbolic communication in Artificial Life simulations. On the other hand, only few works have analyzed the specific evolved communication systems, especially w.r.t. the emerging signals and their association to symbols.

In this work, we present a synthetic model composed of sender and receiver agents controlled by Continuous-Time Recurrent Neural Networks (CTRNNs) \cite{beer_ctrnn_1995}, which are able to generate and decode continuous-time signals. The CTRNNs are optimized by means of the Neuro-Evolution of Augmenting Topologies (NEAT) algorithm \cite{stanley_neat_2002}. First, we show that the agents are able to evolve a signaling system in several settings, even in the presence of noise. Then, we analyze the evolved signaling systems by looking at the properties of the signals evolved under the different conditions. This analysis shows, quite interestingly, that the evolved systems are in most cases similar to well-known signal modulation techniques such as Pulse Amplitude Modulation (PAM) \cite{Barry2004}. Compared to the previous literature, our work is distinct in these two main aspects:
\begin{itemize}[leftmargin=*]
\item We characterize signal decoding as either regression or classification, with limited and unlimited signal amplitude, and show how this choice affects the complexity of the evolutionary search, and leads to different levels of generalization.
\item We perform a thorough analysis of the evolved signaling system in terms of \textit{constellations of signals} \cite{gallager2008principles} and clustering \cite{ankerst1999optics}, and show how different classes of signals emerge from evolution.
\end{itemize}

The rest of the paper is organized as follows. In the next section, we provide an overview of the related work. Then, in \Cref{sec:background} we introduce the background on CTRNNs and NEAT. In \Cref{sec:setup}, we describe the experimental setup. Then, we present the numerical results in \Cref{sec:results}, and we analyze the evolved signaling systems in \Cref{sec:analysis}. Finally, we draw the conclusions of this study in \Cref{sec:conclusions}.


\section{Related work}\label{sec:rw}
Over the years, the evolution of communication (both sub-symbolic and symbolic) has attracted a great research interest from theoretical biologists, mathematicians, linguists, and computer scientists.

Several works analyze the change in communication performance due to the available \textit{implicit} vs. \textit{explicit} communication channels and their properties \cite{goos_learning_2003, cangelosi_evolution_2001, nolfi_evolution_2010, talamini_communication_2019}. One special case of communication based on an implicit channel is \textit{referential communication}, which occurs when agents exchange information by observing the \textit{behavior} of other agents. This form of communication has been largely studied e.g. in \cite{arranz_origins_nodate, shibuya_evolution_2018, williams_evolving_2008, manicka_analysis_2012, ankerst1999optics, evtimova2018emergent, dagan2020coevolution}. 

Social factors, including sexual reproduction \cite{grouchy_evolutionary_2016}, have been shown to play an important role in the evolution of communication. In \cite{smith_cultural_2002}, the author studies the emergence of agreement between symbols and meanings in a ``culture'' (i.e., a population of agents that somehow ``agree'' on a given representation of the reality). In \cite{kalocinski_interactive_2018}, the authors study the effects of social influence and transmission bottlenecks on communication. 

Other works study various specific properties of emerging communication. For instance, Tuci \cite{tuci_investigation_2009} investigate on the ``sound-to-silence'' ratio of an evolved communication system. Lacroix \cite{lacroix_communicative_2020} studies the information transfer in an ``impoverished signaling game''. Froese et al. \cite{froese_modelling_2010}, study how the behavior of the communicating agents changes depending on the scenario. In \cite{lacroix_salience_2020, lacroix_correction_nodate}, Lacroix analyzes the agents' salience, i.e., their expectations w.r.t. the others' behavior. Kadish et al. \cite{kadish_adapting_2020} study the effect of noise on the communication between agents evolved in noise-free contexts. Nowak et al. \cite{nowak_error_1999} study the effect of the number of symbols on the error rate of the language. Talamini et al. \cite{talamini_communication_2020} study how the effectiveness of the communication varies in different settings.



An active research direction in this field focuses on the evolution of \textit{compositional} languages.
A language is defined to be compositional when multiple symbols can be combined to define a meaning. 
For instance in \cite{nolfi_evolving_2010} the authors review the literature related to the emergence of communication and mention the emergence of compositional communication as one of the main future challenges.

Another important aspect of the evolution of communication is the possibility to ``ground'' a language, i.e., create a correspondence between an evolved language and human languages.
In \cite{lazaridou_multi-agent_2017}, the authors train a multi-agent system, composed of a sender and a receiver, to play a referential game. Moreover, they provide a method to ground the language used by NNs in human languages.
Havrylov et al. \cite{havrylov_emergence_2017}, inspired by \cite{lazaridou_multi-agent_2017}, train a communication system based on an encoder-decoder structure, to play a referential game. 
In \cite{chaabouni_anti-efficient_2019}, the authors show that when training a speaker-listener couple of agents, they do not naturally adopt a form of efficient encoding. 
In \cite{harnad_symbol_nodate}, the author describes the symbol grounding problem and proposes a scheme in which he hypothesizes that symbol grounding happens in a bottom-up manner, i.e., starting from perceptions and ending in symbols which describe those perceptions.

While evolutionary approaches are one way to evolve communication systems, they are not the only one. In \cite{barrett_hierarchical_nodate}, the authors study the emergence of communication in a multi-agent system by using simple reinforcement learning in three different versions of a composition game (a kind of referential game). Foerster et al. \cite{foerster_learning_2016} propose two approaches based on Deep Q Networks to make the communication emerge in sender-receiver settings: Reinforced Inter-Agent Learning (RIAL) and Differentiable Inter-Agent Learning (DIAL). In RIAL, sender and receiver learn to communicate by learning independently; in DIAL, the agents learn to communicate by propagating the gradients from receiver to sender.

Finally, the evolution of communication has been object of works tailored to its applications.
In \cite{cambier_language_2020}, the authors present their perspective on the combination of swarm robotics and emergence of language. 
In \cite{kharitonov_egg_2019}, the authors implement a framework to ease the training of multi-agent systems that must learn to communicate.


\section{Background}\label{sec:background}
In the following, we briefly introduce the two main computational tools used in this work, namely CTRNNs and NEAT.

\subsection{CTRNNs}
\label{subsec:ctrnn}
CTRNNs \cite{beer_ctrnn_1995} are computational NNs in which the internal state of each neuron $i$ is time-dependent and it is determined by the following differential equation:
\begin{equation}\label{eq:ctrnn1}
    \tau_i \frac{\partial y_i}{\partial t} = -y_i + f_i\left(\beta_i + \sum\limits_{j \in A_i} w_{ij} y_j\right)
\end{equation}
where $\tau_i$, $y_i$, $f_i$ and $\beta_i$ indicate respectively the time constant, the potential, the activation function and the bias of neuron $i$, $A_i$ is the set of indices of neurons that provide an input to neuron $i$, and $w_{ij}$ is the weight of the connection from neuron $j$ to neuron $i$.

The time evolution of a CTRNN can be computed using the forward Euler method:
\begin{equation}\label{eq:ctrnn2}
    y_i(t+\Delta t) = y_i(t) + \Delta t \frac{\partial y_i}{\partial t}
\end{equation}

One interesting property of CTRNNs is that they inherently allow recurrence and self-loops. Because of this, complex internal patterns can emerge. The resulting temporal dynamics makes CTRNNs particularly suited to model systems that are capable to produce an output signal \cite{bown_ctrnn_2006} or to receive an input signal.


\subsection{NEAT}
\label{subsec:neat}
Neuro-evolution, that is the application of EAs to optimize NNs, is a growing field in computational intelligence \cite{stanley2019designing}. While seminal works in this direction were oriented toward the optimization of NNs with fixed topology, in the last two decades methods have been proposed to evolve the topology along with the weights.

One of the main algorithms falling into this category is NEAT \cite{stanley_neat_2002}. NEAT combines tuning the network weights with \textit{complexification}, meaning that the algorithm also evolves the network topology to fit the complexity of a given problem. Usually, NEAT starts with a population made of the simplest possible NNs, and changes in the network topology are introduced by \textit{structural mutations} that can add or remove edges and nodes. This process allows NEAT to select, through fitness evaluations, the right topology with an appropriate level of complexity with respect to the problem to solve. On a very high level, to allow structural mutations, each gene in each genome refers to an edge in the NN, and has information about the two nodes being connected and the weight of that connection.

NEAT speciates the population based on genomes' similarity and historical ancestors (\textit{historical markings}). In this way, individuals are constrained to compete primarily within their own niches (\textit{explicit fitness sharing}), and then with the whole population. This way of proceeding ensures topological innovations to be protected enough and have time to optimize their structure before having to compete with other species in the population. For a more in-depth description of NEAT, we refer to the reader to \cite{stanley_neat_2002}.


\section{Experimental setup}\label{sec:setup}
Before diving into the description of our experimental setup, we provide some definitions that we will use in the rest of the text.

\begin{definition}[Concept]
A concept is broadly defined as an idea or a concrete object that a population of agents is interested in being able to communicate.
\end{definition}

\begin{definition}[Symbol]
A symbol is defined as a social agreement on the representation of a concept. In our case, the representation is a $1$-D signal, and we assume a 1-to-1 concept-symbol mapping.
\end{definition}

\begin{definition}[Vocabulary]
A vocabulary, denoted as $V$, is defined as the collection of concepts shared (or intended to be shared) by a population of agents, and their associated symbols.
\end{definition}

The general setting of the problem of our interest is the following. We consider two agents, one called the \textit{sender}, and the other one called the \textit{receiver}. Both agents share a common set of concepts, and our goal is to evolve an agreement on a signal representation for all concepts (i.e., agents need to create a common vocabulary). Once a common vocabulary is established, the sender is able to propagate information to the receiver (this might be used, e.g., to communicate to the receiver what to pick in a collection of objects, or to indicate in which direction to move in a navigation task).

In the first part of our experiments, concepts are represented by one feature (\cref{sec:symbolic_comm}-\ref{sec:noise_comm}), i.e., the ``id'' of the concept (represented by a number). Later, we also investigate the case in which concepts are represented by $m$ features, introducing a referential game \cite{lazaridou_multi-agent_2017} (\cref{sec:referential}). In all cases, we focus on the effect of two specific aspects on the evolved signal representation:
\begin{itemize}[leftmargin=*]
    \item The \textbf{decoding} paradigm, i.e., whether modeling signal decoding as a \emph{regression} or a \emph{classification} problem has any influence on the evolutionary process. In the regression setting, the receiver's NN has only one output node, i.e., the concept it decoded. In the classification setting, the receiver's NN has $|V|$ output nodes (see \cref{spec-table}), representing the probability distribution over the vocabulary $V$. While this latter approach is more robust (as it allows probabilistic decoding, thus being potentially less affected by noise), regression is interesting in that it is associable to numerical meaning of counting, one of the main concepts that humans acquire in the first 3-4 years of childhood. This, in turn, may facilitate the acquisition of higher-level concepts \cite{carey2000origin}.
    \item The available \textbf{amplitude} of the signals generated by senders, which can be \emph{limited} or \emph{unlimited} depending on the sender activation function (see \cref{spec-table}). Intuitively, evolving a symbolic representation for a set of concepts in a limited amplitude setting is much more difficult with respect to an unlimited setting, since signal representations need to compete for a limited bandwidth and avoid overlapping, which may lead to ambiguity.
\end{itemize}
Thus, we consider four settings in total, that we refer to \emph{regression limited/unlimited} and \emph{classification limited/unlimited} respectively. For each experimental setting under investigation, we are interested in analyzing the evolution of the agreement of the signal representation in the population of senders and receivers.

\subsection{Agents}
A sender-receiver pair can be seen as an encoder-decoder structure. More specifically, the \textit{sender} is controlled by a CTRNN that takes as input the representation of an object (that may be either its id or a set of properties, as shown in Section \ref{sec:referential}) and sends to a channel a continuous-time signal (which is sampled into a fixed number of equally-spaced samples over a given time window) that encodes a symbol associated to that object. On the other hand, the \textit{receiver} is controlled by a CTRNN that takes as input the sampled signal sent by the sender and has to \textit{decode} it into the corresponding object. With reference to Eq. (\ref{eq:ctrnn2}), we assume $\Delta t=1$. We should node that this configuration is purposely supposed to ``force'' the communication between the agents: in fact, by driving the evolutionary process towards the communication between the agents, we expect to observe the emergence of a symbolic communication system.


\subsection{Co-evolutionary algorithm}
\label{sec:algo}
In order to evolve the CTRNNs, we used the implementation of NEAT provided by the \texttt{NEAT-Python} library\footnote{Available at: \url{https://github.com/CodeReclaimers/neat-python}.}, and built on top of it a co-evolutionary algorithm to allow the evolution of sender/receiver pairs\footnote{Our code is publicly available at: \url{https://github.com/FraLotito/evol-signal-comm}.}. Tables \ref{comm-table}-\ref{spec-table} show the main parameters used in the experiments, which are based on the default values set in \texttt{NEAT-Python}.

The co-evolutionary algorithm (whose full pseudo-code is omitted for brevity) consists of two parallel independent executions of NEAT, one handling a population of senders and one handling a population of receivers, which are synchronized at each generation on the evaluation process. The latter, whose pseudo-code is shown in \cref{alg:co-ev}, consists in the following. Sender/receiver pairs are formed by the Cartesian product between all sender genomes and receiver genomes. Each pair $p$ uses it time window to propagate the signal (from sender to receiver), after which the fitness of $p$ (shared between sender and receiver), to be maximized, is computed as:

\begin{equation}\label{fitness_equation}
    \textit{p.fitness} = - \sum_{\textit{concepts}} \mathbb{1}[concept_{sent} \neq concept_{received}]
\end{equation}
\noindent{}where \textit{concepts} indicates the set of concepts considered for the evaluation of the pair, and $\mathbb{1}[concept_{sent} \neq concept_{received}]$ is an indicator function returning 1 if the concept sent is different from the concept received and 0 otherwise.

Since each single sender/receiver genome is part of multiple pairs, eventually we consider as its final fitness the best one (maximum) among the fitnesses of all the pairs to which it participated.

The co-evolutionary process stops either when a pair is found with fitness equal to zero (indicating no communication error), or after a total budget of 10000 generations for each of the two populations. Each population is randomly reset in case of 50 generations without improvement.

\begin{algorithm}[ht!]
\SetAlgoLined
 Let $S, R$ be the sets of sender and receiver genomes\;
 Let $P = S \times R$ be the set of sender/receiver pairs to evaluate\;
 Let $x.\textit{fitnesses}$ be the set of fitnesses computed $\forall x \in S, R$\;
 Let \textrm{computeFitness()} be the function reported in Eq. \eqref{fitness_equation}\; 
 \ForEach{$p \in P$}{
  $\textit{p.fitness} = \textrm{computeFitness}(p)$\;
  $\textit{p.sender.fitnesses} = \textit{p.sender.fitnesses} \cup \textit{p.fitness}$\;
  $\textit{p.receiver.fitnesses} = \textit{p.receiver.fitnesses} \cup \textit{p.fitness}$\;
 }
 \ForEach{$x \in S, R$}{
 $x.\textit{fitness} = \max{(x.\textit{fitnesses})}$\;
 }
 \caption{Co-evolutionary evaluation process.}
 \label{alg:co-ev}
\end{algorithm}

\begin{table}[ht!]
\centering
\caption{Common parameter settings.}
\resizebox{0.45\textwidth}{!}{
\begin{tabular}{|l|c|l|c|}
 \hline
 \textbf{Parameter} & \textbf{Value} & \textbf{Parameter} & \textbf{Value}\\
 \hline
 Sender population & 20 &
 Receiver population & 20\\
 Time window & 10 &
 Sender input nodes & 1\\
 Sender hidden nodes & 0 &
 Sender output nodes & 1\\
 Receiver input nodes & 1&
 Receiver hidden nodes & 0\\
 Connection add prob. & 0.5 & Connection delete prob. & 0.5\\
 Node add prob. & 0.2 & Node delete prob. & 0.2\\
 Elitism (species) & 1 & Elitism (individual) & 1\\
 Reset on extinction & 1&
 Max generations (stagnation) & 50\\
 Feed forward & False&
 Max generations (total) & 10000\\
 \hline
\end{tabular}
}
\label{comm-table}
\end{table}

\begin{table}[ht!]
\centering
\caption{Specific parameter settings.}
\resizebox{\linewidth}{!}{
\begin{tabular}{|l|c|c|c|c|}
 \hline
 \multirow{2}{*}{\textbf{Name}} & \multicolumn{2}{c|}{\textbf{Regression}} & \multicolumn{2}{c|}{\textbf{Classification}}\\
 \cline{2-5}
 & \textbf{Unlimited} & \textbf{Limited} & \textbf{Unlimited} & \textbf{Limited}\\
 \hline
 No. receiver outputs & 1 & 1 & $|V|$ & $|V|$\\
 \hline
 Sender activation func. & Identity & Sigmoid & Identity & Sigmoid\\
 \hline
 \multirow{2}{*}{Receiver activation func.} & \multirow{2}{*}{Identity} & \multirow{2}{*}{Identity} & Sigmoid & Sigmoid\\
 & & & + Softmax & + Softmax\\
 \hline
\end{tabular}
}
\label{spec-table}
\end{table}


\section{Results}\label{sec:results}
We present now the results of our experiments: first we consider simple symbolic communication, then we test zero-shot communication, and communication in the presence of noise. Finally, we apply our evolutionary model to a multi-feature referential game.

\subsection{Symbolic communication}
\label{sec:symbolic_comm}
The first question we try to address is whether the evolution of symbolic communication is facilitated or not by: 1) the use of a limited or unlimited signal amplitude and 2) formulating the decoding step as either a regression or classification problem.

In order to answer this question, we consider the four aforementioned settings, and evolve agents to share a collection of $|V|$ concepts (i.e., a vocabulary of size $|V|$), each one described by one feature. We assume that there exists an order (shared in the population) between the concepts, such that each concept is correctly identified by its index. The concept $i$ is described by the feature $\frac{i}{|V|}$, therefore the feature representation is limited in the interval $[0,1]$.

\begin{figure*}[ht!]
\centering 
\includegraphics[width=.9\linewidth,trim={0cm 1.5cm 0 1cm},clip]{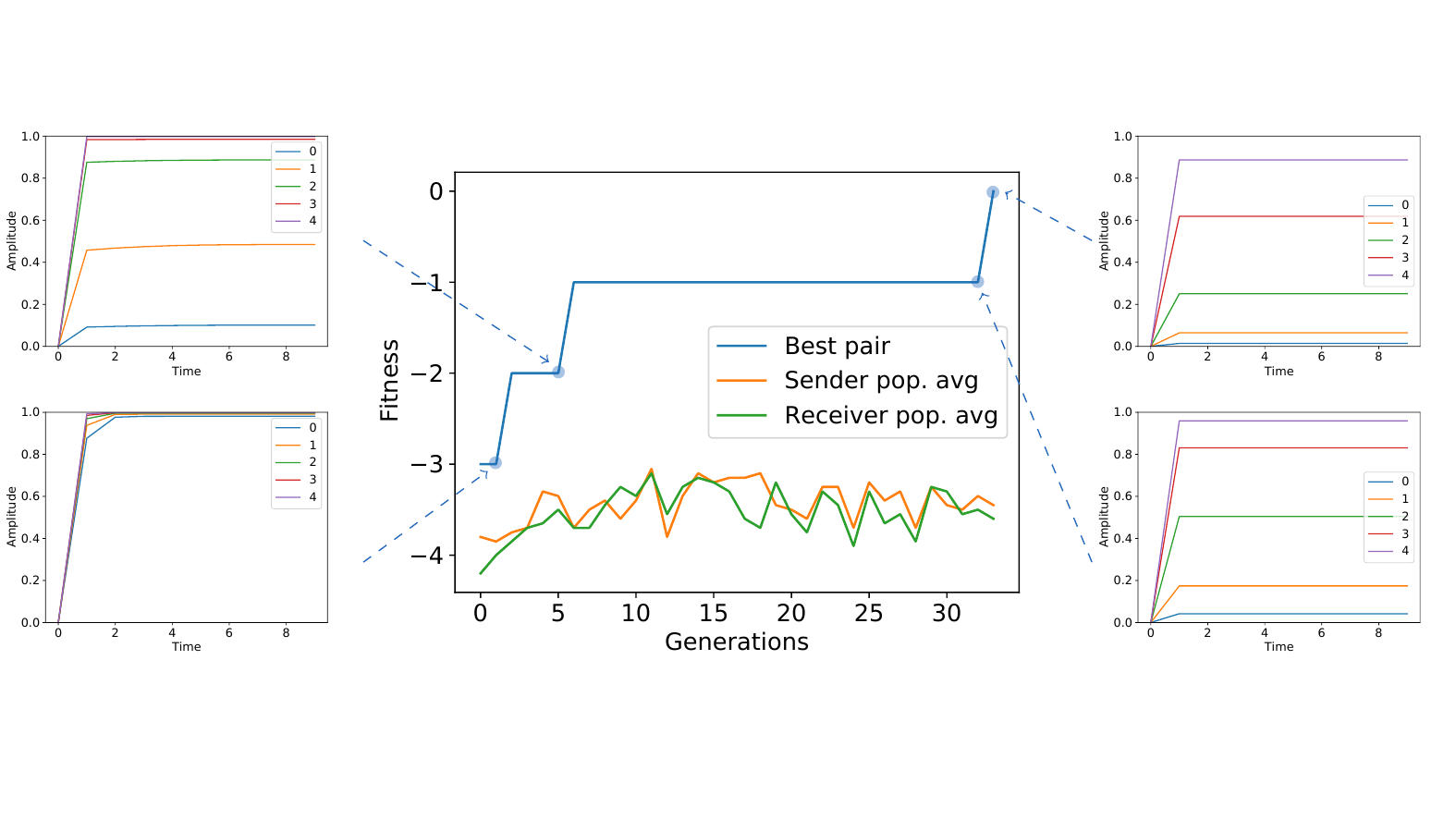}
\caption{Evolution of signals across generations in the regression setting with limited amplitude. The agents first evolve a coarse subdivision of the amplitude, and then progressively refine it, until all concepts are correctly communicated.}
\label{fig:zoom}
\end{figure*}

\Cref{fig:zoom} shows the evolutionary trend an example run on the regression setting with limited amplitude, with the corresponding signals evolved at different steps of the process. In the first generations, the subdivision of the amplitude between the different signals is very coarse, leading to ambiguity in the symbolic representation, and therefore to low fitness value. The evolution eventually leads to a change in the shape of the signals (from triangular to constant) and to an unambiguous subdivision of the amplitude.

\begin{figure}[ht!]
  \centering
  \subfloat[Regression]{\label{fig:reg_complexity}\includegraphics[width=0.24\textwidth]{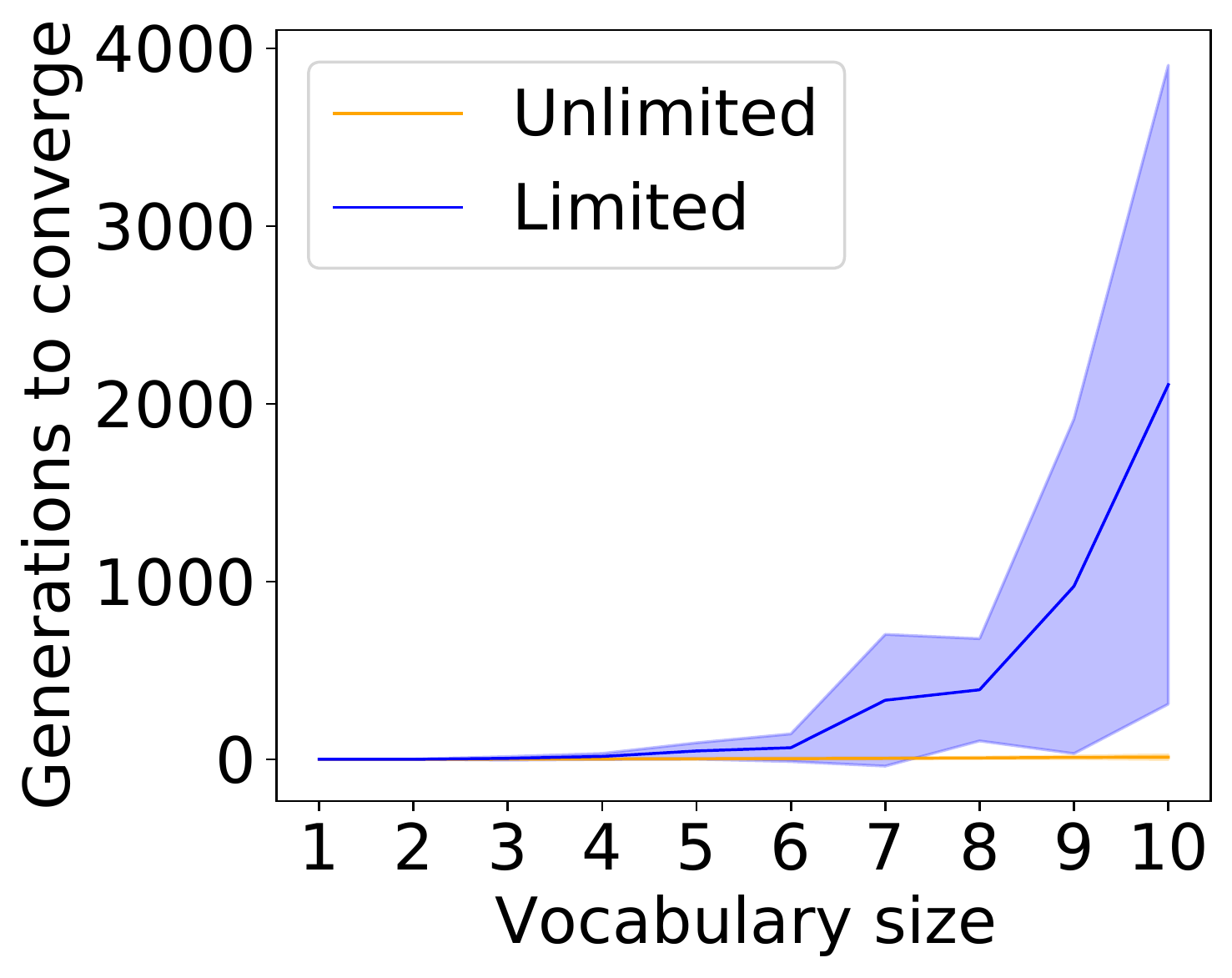}}
  \subfloat[Classification]{\label{fig:class_complexity}\includegraphics[width=0.24\textwidth]{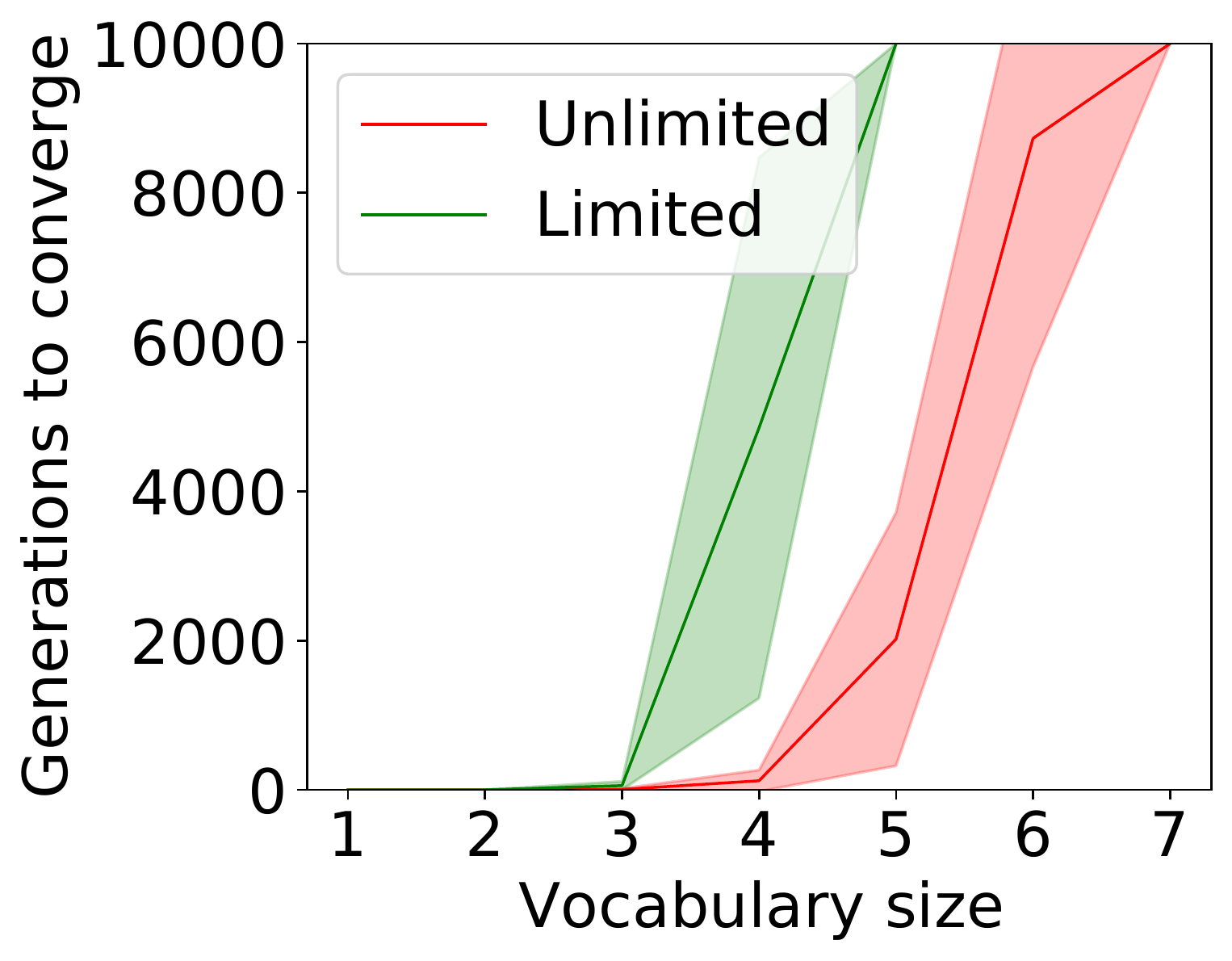}}
  \caption{Number of generations to converge in the four different settings. Note that the two subplots are represented with different x-y scales.}
  \label{fig:complexity}
  \vspace{-0.5cm}
\end{figure}

\begin{figure}[ht!]
  \centering
  \subfloat[Regression, unlimited]{\label{fig:trend_reg_unl_5}\includegraphics[width=0.24\textwidth]{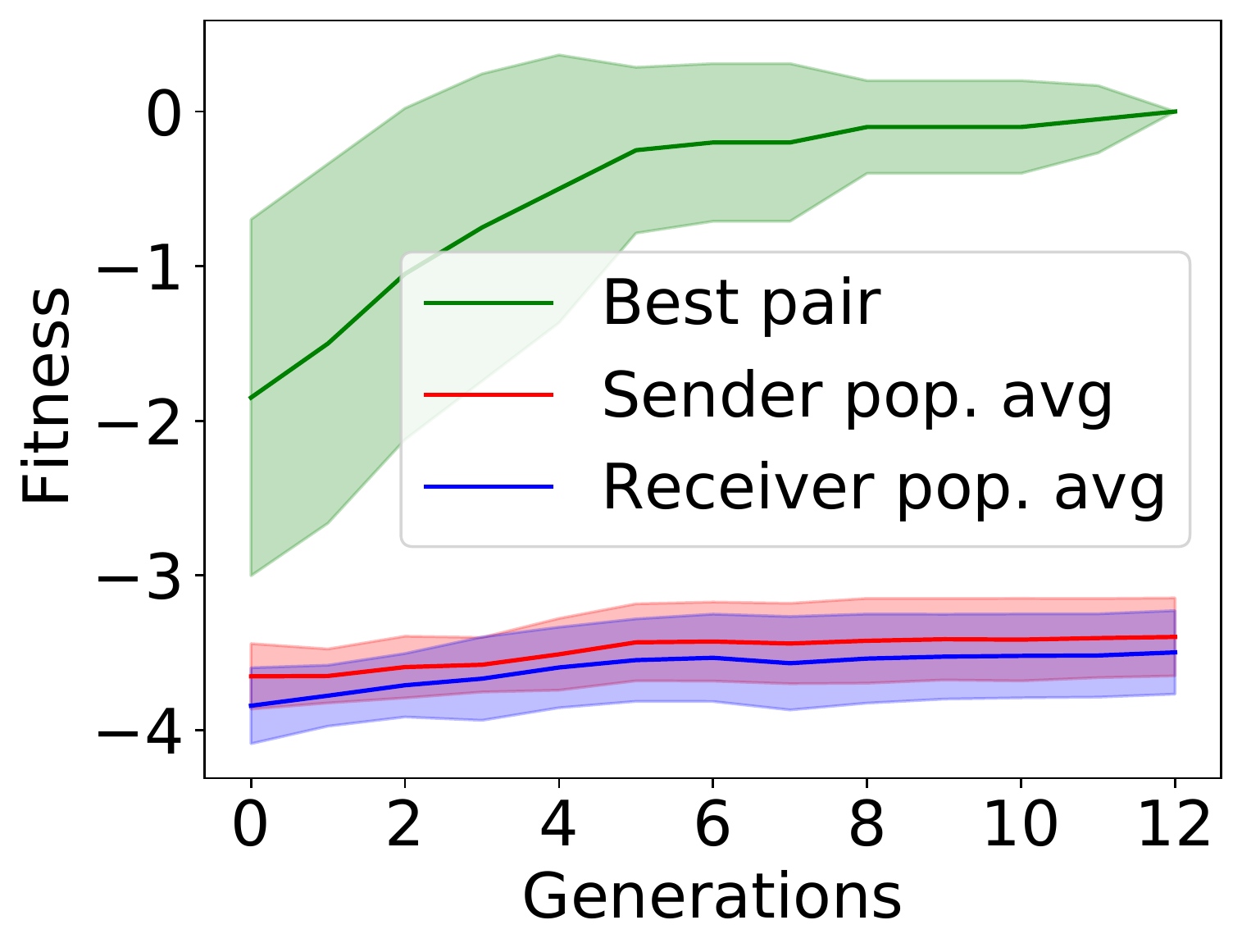}}
  \subfloat[Regression, limited]{\label{fig:trend_reg_lim_5}\includegraphics[width=0.24\textwidth]{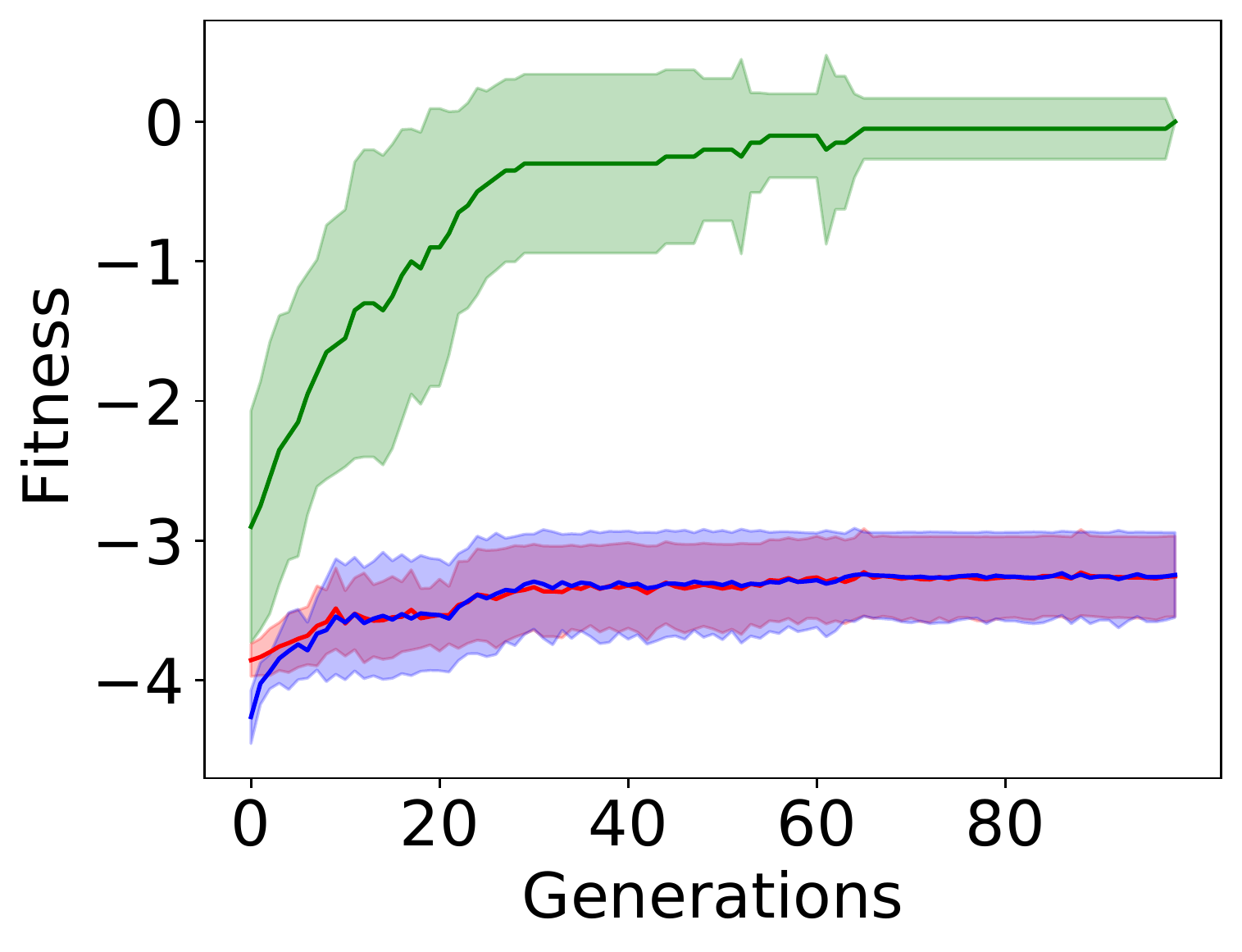}}

  \subfloat[Classification, unlimited]{\label{fig:trend_class_unl_5}\includegraphics[width=0.24\textwidth]{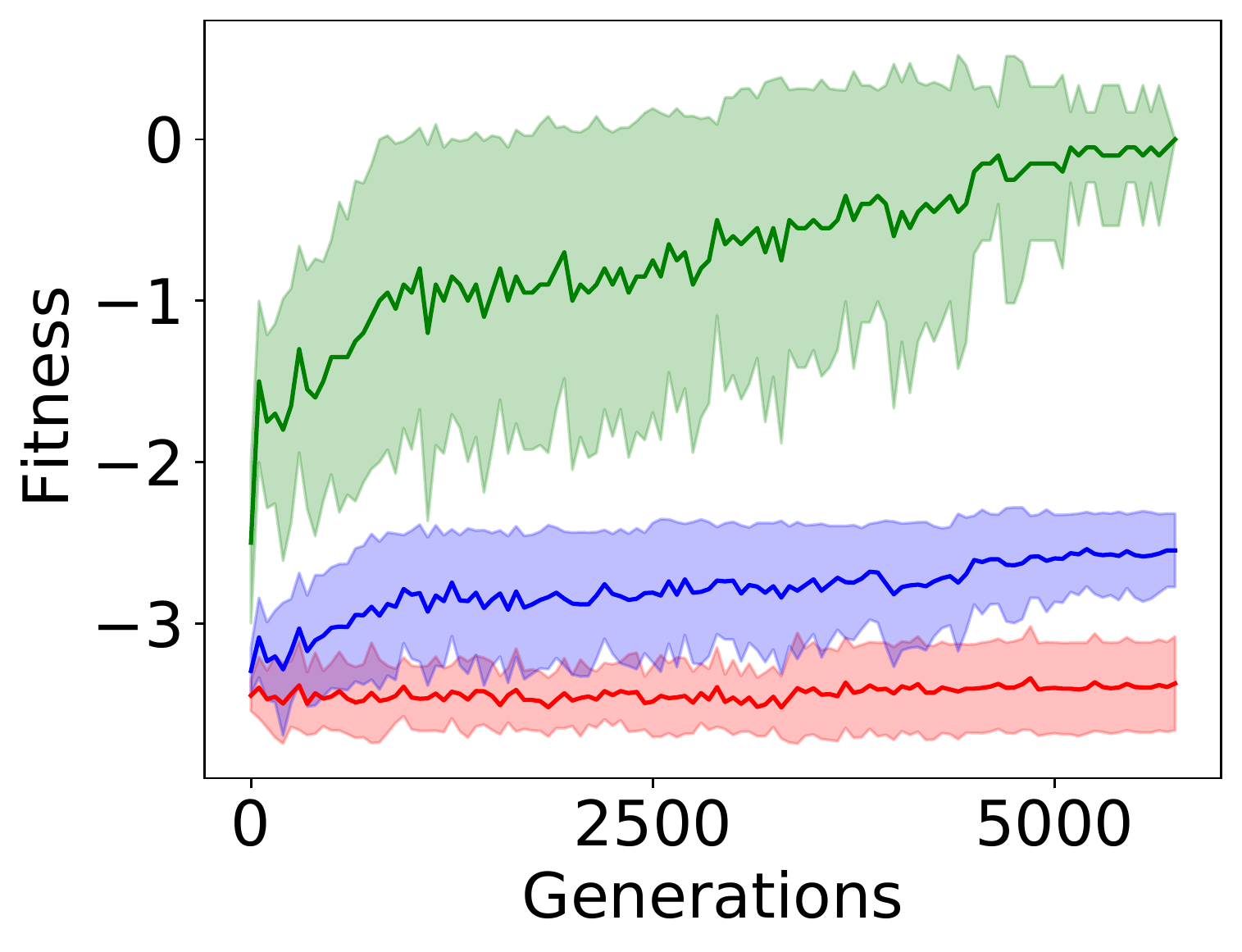}}
  \subfloat[Classification, limited]{\label{fig:trend_class_lim_4}\includegraphics[width=0.24\textwidth]{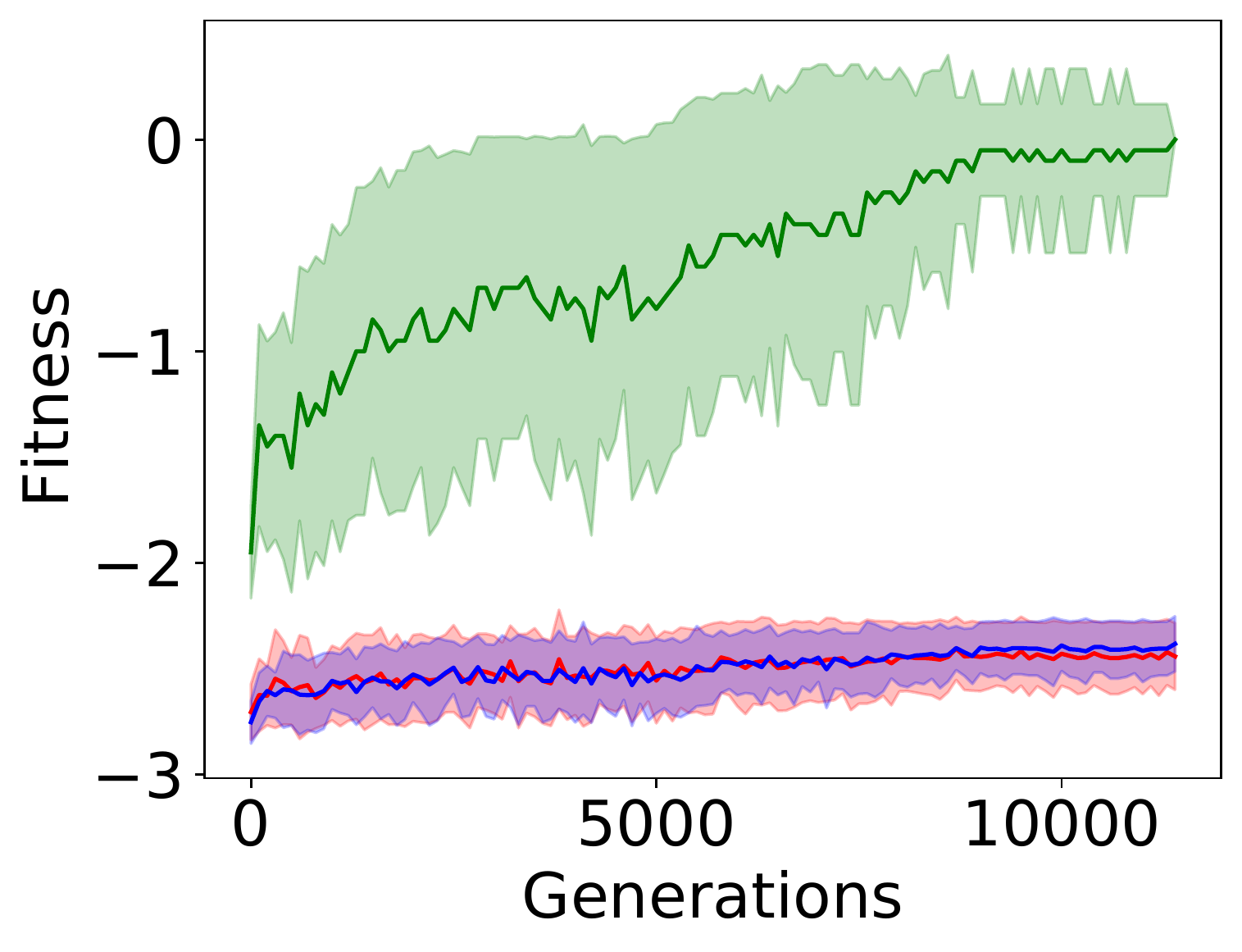}}
  \caption{\label{fig:trends}Fitness trend (averaged over 20 evolutionary runs) in the four different settings with a vocabulary of size 5 (4 for the classification setting with limited amplitude).}
  \vspace{-0.5cm}
\end{figure}


\Cref{fig:complexity} shows the number of generations required for the evolution to converge to a zero-fitness solution, as a function of the vocabulary size. For each vocabulary size, the corresponding value is averaged over $20$ evolutionary runs. We observe that there is a large gap in the convergence time (which can be seen as a proxy to complexity of the problem) when comparing the regression and classification settings, the first one being easier to solve than the latter. We further observe that limiting the available amplitude for the signals increases the hardness of the problem in both settings.

\Cref{fig:trends} shows the fitness trend (averaged over $20$ runs) in each of the four settings, with a vocabulary size $|V|=5$ ($4$ for the classification setting with limited amplitude, for which it was not possible to converge to a solution within the 10000 generations, see \cref{fig:class_complexity}). Also in this case we note that the regression settings are quickly solved in a few tens of generations, while in the classification settings the evolution shows a much slower convergence.

\begin{table}[ht!]
    \centering
    \caption{Zero-shot experimental results (no. of correctly communicated concepts, including $|V_T|$).}
    \resizebox{\linewidth}{!}{
    \begin{tabular}{|l|c|c|c|}
        \hline
        \textbf{Setting} & \textbf{$|V_T| = 3$} & \textbf{$|V_T| = 5$} & \textbf{$|V_T| = 7$} \\
       \hline
        Regression, unlimited & $4.75 \pm 2.4$ & $7.90 \pm 1.7$ & $9.20 \pm 1.2$ \\
        Regression, limited & $3.15 \pm 0.5$ & $5.10 \pm 0.3$ & $7.30 \pm 0.5$ \\
        Classification, unlimited & $3.15 \pm 0.4$ & $5.00 \pm 0.0$ & NA \\
        Classification, limited & $3.00 \pm 0.0$ & NA & NA \\
    \hline
    \end{tabular}
    }
    \label{tab:generalization}
\end{table}

\subsection{Zero-shot communication}
\label{sec:zeroshot_comm}
The second question we address is whether the evolved pairs of agents under the four settings are capable of performing zero-shot communication, i.e., correctly communicate concepts not seen at fitness evaluation time. In other words, to assign an unseen concept to the sender and see if the sender/receiver communication is correctly carried out, even if there was no previous agreement on the signal to use. More formally: let $V$ be the vocabulary and $V_T \subset V$ be the set of concepts considered for fitness evaluation, we are interested in observing if the agents in the best evolved pair $p$ are able to communicate, at the end of the evolutionary process, also the concepts in $V - V_T$. 

In these experiments, we set $|V| = 10$, and we measure the number of concepts in $V$ (including those in $|V_T|$) that are correctly communicated by the best pair at the end of the evolution, for three different values of $|V_T|$. The results are reported in \Cref{tab:generalization}, averaged over $20$ evolutionary runs. In the table, ``NA'' means that the setting did not converge to a solution. We observe that zero-shot communication occurs effectively only in the regression cases with unlimited amplitude, where 2 to 3 unseen concepts are communicated correctly. On the contrary, the classification settings seem to be unable to generalize to unseen concepts.

\subsection{Communication in the presence of noise}
\label{sec:noise_comm}
So far we have assumed the channel of communication to be reliable, a strong assumption considering that many real-world channels are actually subject to noise. In order to investigate the effect of noise, we replicate the experiments of \cref{sec:symbolic_comm}, and apply Gaussian noise $\mathcal{N} \sim (0, \sigma^2)$ over the signals generated by senders. Also in this case we set $|V|=5$ for all cases except the classification setting with limited amplitude, for which we set $|V|=4$.

We devised two different sets of experiments. In the first one (results reported in \cref{tab:noise1}), the fitness of each sender/receiver pair is computed as in Eq. \eqref{fitness_equation}, with the sender trying to communicate each concept in $V$ once. In the second one (results reported in \cref{tab:noise3}), during the fitness evaluation each concept is communicated $3$ times (i.e., there are $3 \times |V|$ trials of communication), and Eq. \eqref{fitness_equation} is modified accordingly to account for the $3$ trials per concept. In this second experiment, the convergence is expected to be more difficult than in the first one, since a sender/receiver pair needs to be able to succeed in more communication trials subject to noise, thus requiring to evolve a much more robust communication system.

After evolution, we then consider a test phase during which the best evolved pair performs, for each concept in $V$, $20$ communication trials ($25$ times for $|V|=4$), thus for a total of $100$ test cases for each pair. To control the amount of noise in the system, we consider four different values for $\sigma$, applied both at evolution and test time.

In \cref{tab:noise1,tab:noise3} we report, for each setting, the results in terms of successfully communicated concepts over $100$ test cases, averaged over $20$ runs. In the tables, ``NA'' means that the setting did not converge to a zero-fitness solution. Overall, the percentage of successful communication tends to decrease as the noise level increases. A notable exception to this trend is the regression unlimited amplitude case with $\sigma = 1$ and $3$ trials per concept (\cref{tab:noise3}), where the percentage of success increases w.r.t. $\sigma = 0.5$ . This effect seems to suggest that there is a certain noise threshold after which large noise levels force signals to be better separated (especially in the unlimited amplitude case where there is no constraint on the inter-signal distance), thus leading to lower ambiguity/higher robustness and a higher percentage of successful communication. Although, the same effect is not observed in the corresponding classification case. This intuition is however confirmed by observing the equally spaced constellations of signals corresponding to this case, reported in the Supplementary Material (available on our GitHub repository).

Another interesting finding is that in the unlimited amplitude cases the classification setting consistently outperforms the regression setting in terms of robustness to noise. Finally, we observe that the setting with $3$ trials per concept is more challenging since there are cases in which the evolutionary process does not converge to a solution; however, when convergence occurs this setting leads to higher percentages of success w.r.t. the corresponding cases with $1$ trial per concept.

\begin{table}[ht!]
    \centering
    \caption{Noise experimental results (1 trial per concept).}
    \resizebox{\linewidth}{!}{
    \begin{tabular}{|l|c|c|c|c|}
        \hline
        \textbf{Setting} & \textbf{$\sigma = 0.1$} & \textbf{$\sigma = 0.2$} & \textbf{$\sigma = 0.5$} & \textbf{$\sigma = 1$} \\
        \hline
        Regression, unlimited & $93.00 \pm 8.60$ & $72.40 \pm 19.5$ & $52.35 \pm 30.2$ & $37.15 \pm 27.2$\\
        Regression, limited & $62.70 \pm 8.30$ & $36.90 \pm 16.1$ & $18.45 \pm 3.30$ & $17.50 \pm 3.90$\\
        Classification, unlimited & $96.55 \pm 9.70$ & $94.55 \pm 7.20$ & $88.25 \pm 9.30$ & $72.75 \pm 29.2$\\
        Classification, limited & $47.45 \pm 18.2$ & $28.25 \pm 9.90$ & $24.50 \pm 6.30$ & $26.40 \pm 4.60$\\
        \hline
    \end{tabular}
    }
    \label{tab:noise1}
\end{table}

\begin{table}[ht!]
    \centering
    \caption{Noise experimental results (3 trials per concept).}
    \resizebox{\linewidth}{!}{
    \begin{tabular}{|l|c|c|c|c|}
        \hline
        \textbf{Setting} & \textbf{$\sigma = 0.1$} & \textbf{$\sigma = 0.2$} & \textbf{$\sigma = 0.5$} & \textbf{$\sigma = 1$} \\
        \hline
        Regression, unlimited & $95.90 \pm 4.6$ & $88.60 \pm 8.00$  & $82.60 \pm 9.3$ & $94.45 \pm 8.9$\\
        Regression, limited & $67.45 \pm 5.1$ & $65.65 \pm 14.0$ & NA & NA\\
        Classification, unlimited & $97.00 \pm 5.5$ & $96.50 \pm 4.00$ & $93.35 \pm 4.6$ & $90.95 \pm 5.3$\\
        Classification, limited & $74.70 \pm 7.6$ & NA & NA & NA\\
        \hline
    \end{tabular}
    }
    \label{tab:noise3}
\end{table}

\subsection{Referential game}\label{sec:referential}
\begin{figure}[ht!]
    \centering
    \includegraphics[width=\linewidth]{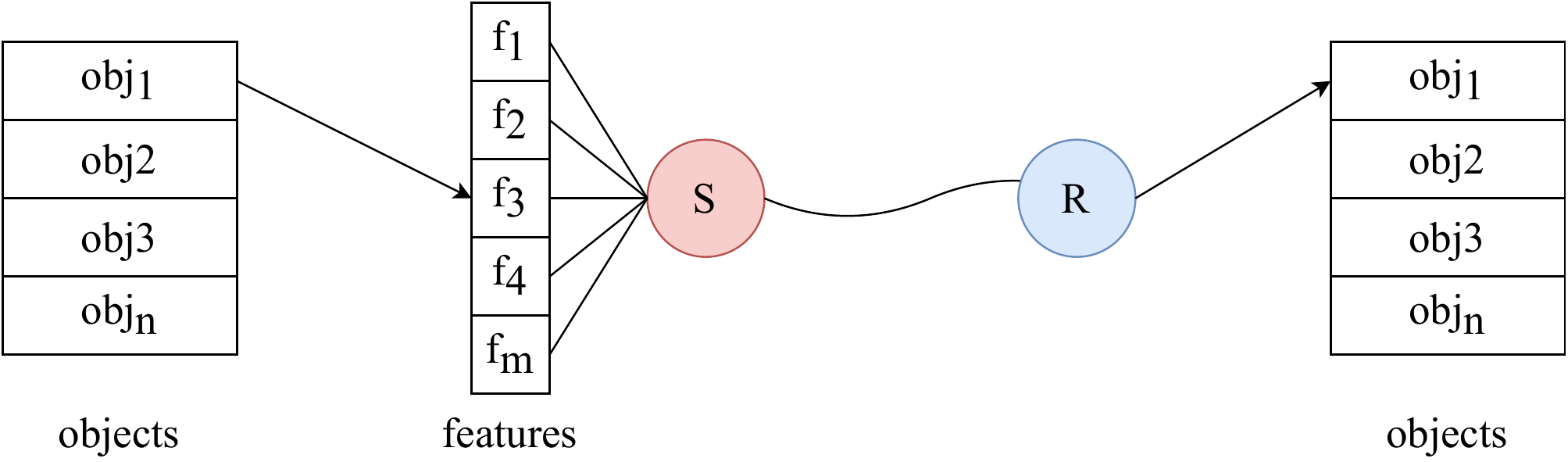}
    \caption{Conceptual scheme of the referential game.}\label{fig:referential}
\end{figure}

To further analyze the capabilities of the evolved communication system, we introduce a simple referential game \cite{lazaridou_multi-agent_2017}. Both sender and receiver are provided with a collection of $N$ objects, each made up of $m$ binary features. The order of the objects is preserved in both collections, such that each object is identifiable by its position without ambiguity. The sender chooses an object, and, after a time window allotted for the communication, the receiver is requested to pick the same object chosen by the sender, see \cref{fig:referential}.

In this game, the sender/receiver pair is required to evolve an effective communication protocol based on $1$-D signals which describe $m$-dimensional feature representations of objects.

We rely again on the co-evolutionary algorithm described in \cref{sec:algo}, computing the fitness of each sender/receiver pair $p$ (to be maximized) as follows:
\begin{equation}\label{fitness_multivariate}
    \textit{p.fitness} = - \sum_{\textit{objects}} \mathbb{1}[object.id_{sender} \neq object.id_{receiver}]
\end{equation}
\noindent{}where:
\begin{itemize}[leftmargin=*]
    \item \textit{objects} represents the set of objects available in the experiment;
    \item $\mathbb{1}[object.id_{sender} \neq object.id_{receiver}]$ is an indicator function returning 1 if the object picked by the sender is different from the object picked by the receiver and 0 otherwise.
\end{itemize}
Therefore, to compute the fitness of each pair, the referential game is repeated once for every object in the set, and the fitness considers the number of objects correctly recognized. Also in this case we take as final fitness of each agent the maximum among the fitnesses of all the pairs to which it participated. The stop criteria and all the other parameters are set as \cref{sec:algo}. In this case the optimal solution is to have a communication protocol that allows the receiver to recognize every object picked by the sender.

Each object is described by $3$ binary features, yielding a maximum of $8$ different objects. We perform experiments with $3$ different sizes of the set of the objects to be shared ($3$, $5$ and $8$). In the case of  $3$ and $5$ objects, these are chosen randomly (once for all the runs of each setting) among the $8$ possible triples of bits. For each size and setting, we execute $20$ evolutionary runs, and average the results. 

\begin{figure}[ht!]
  \centering
  \subfloat[Regression]{\label{fig:trend_reg_multi}\includegraphics[width=0.24\textwidth]{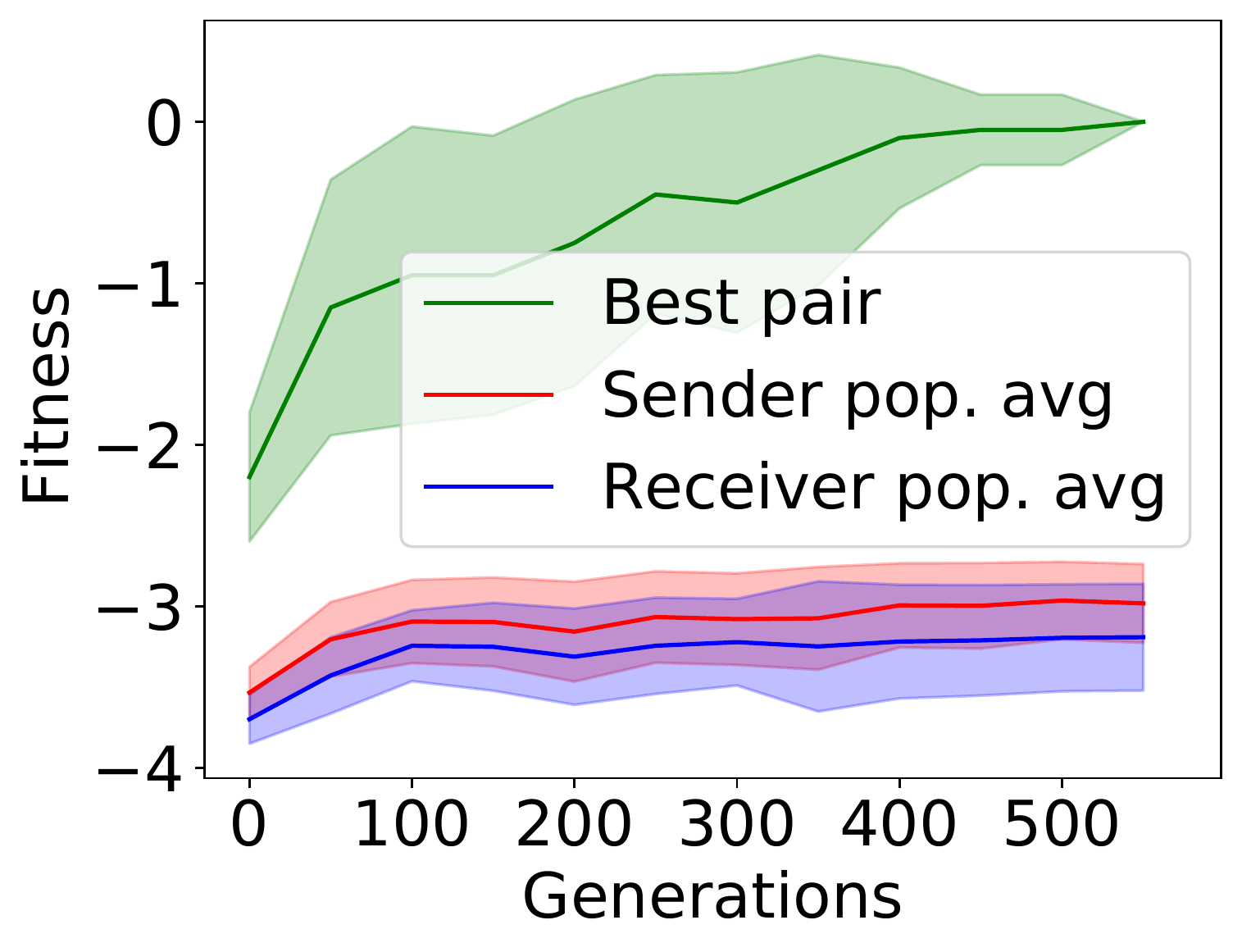}}
  \subfloat[Classification]{\label{fig:trend_class_multi}\includegraphics[width=0.24\textwidth]{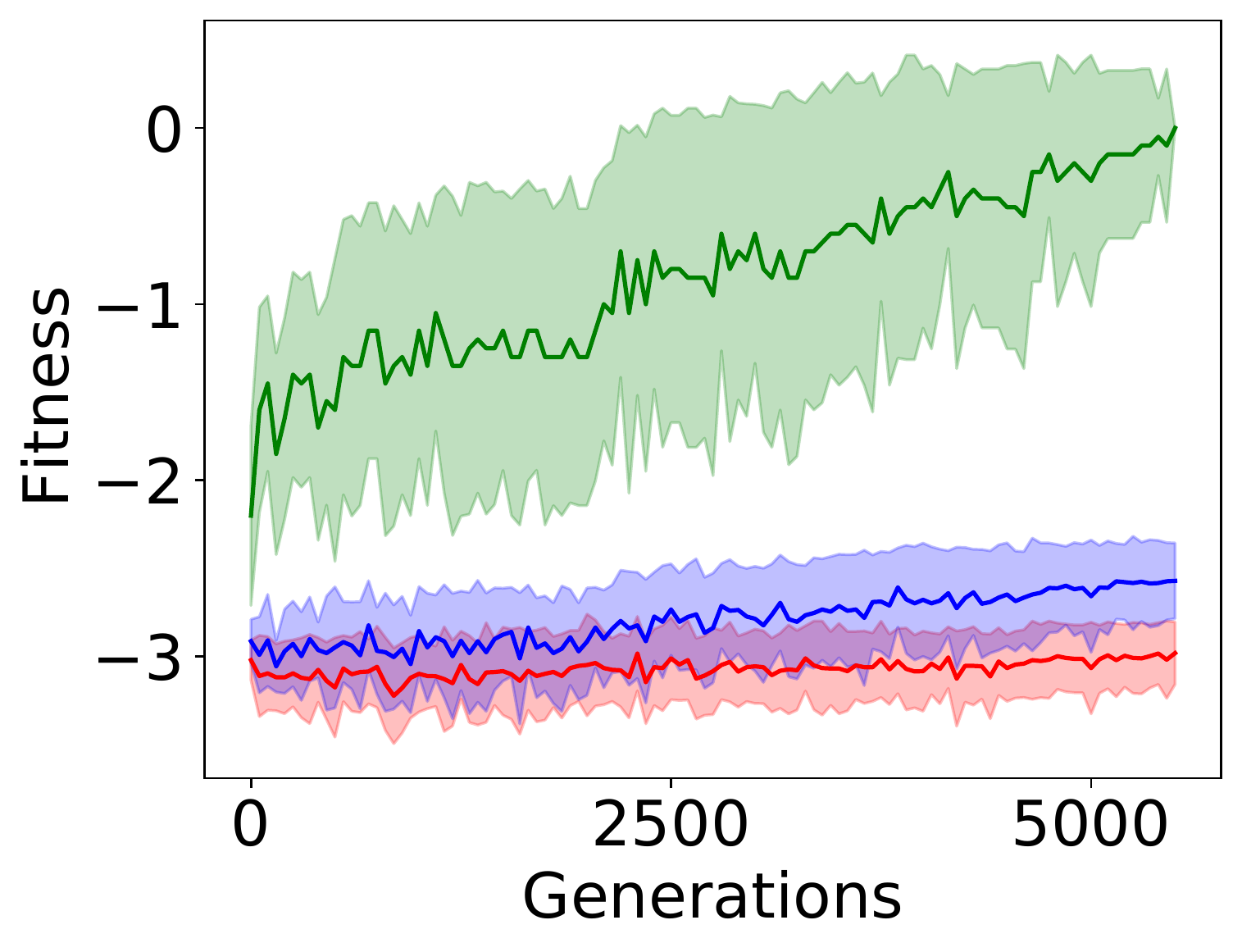}}
  \caption{\label{fig:trend_ref} Fitness trend of the referential game experiments in the regression and classification settings with unlimited amplitude and a vocabulary made up of 5 objects.}
\end{figure}

Considering the unlimited amplitude case, both under regression and classification settings, the evolution achieves $100\%$ success rate in the case of $3$ and $5$ objects, while it fails to converge to a solution in the case of $8$ objects. For illustration purposes, we report the fitness trend regarding the regression and classification settings with $5$ objects in \cref{fig:trend_ref}, where once again we can notice the slower convergence of the classification setting.

In the experiments with limited amplitude, both the regression and classification settings fail to reach convergence for any number of objects from $3$ to $8$, confirming our previous observations on the increased difficulty due to the use of limited signal amplitudes.

\section{Analysis of the evolved signaling systems}
\label{sec:analysis}

As seen earlier, the evolved signaling systems can be seen as an agreement between two parties (i.e., the sender and the receiver) that must optimize their behavior in order to find an optimal encoding-decoding strategy: this, in turn, can be seen as a trade-off between simplicity (for the receiver to understand the meaning of signals) and expressiveness (for the sender to convey the meaning of signals). In this section, we will analyze the communication systems evolved, and we will show how the signals evolved tend to form clusters characterized by different specific waveforms.

\subsection{Constellations}
In order to understand how the signaling systems evolved, we apply the procedure described in \cite{gallager2008principles} to analyze the constellations of signals produced by the system in the regression setting with limited amplitude, $|V|=5$ and without noise. In our model, the signals $s_k$, $k=1\dots|V|$, represent time series of $10$ samples each (size of the time window). The procedure from \cite{gallager2008principles} works as follows. In the first step, we take the first signal from the evolved signaling system, $s_1$, and compute $\varphi_1$, i.e., the first base of a vector space, as:
\[ \varphi_1 = \frac{s_1}{\|s_1\|}\]
where $\|\cdot\|$ indicates the 2-norm. Then, we iterate on each signal $s_k$, $k=2\dots|V|$, and compute $(s_{k})_{\bot (k-1)}$, that is its orthogonal component with respect to all the previous bases:
\[ (s_{k})_{\bot (k-1)} = s_{k} - \sum\limits_{i=0}^{k-1} \langle s_{k}, \varphi_i \rangle \varphi_i\]
where $\langle\cdot,\cdot\rangle$ indicates the scalar product. With this, we calculate a new base $\varphi_{k}$ as follows:
\[ \varphi_k = \frac{(s_{k})_{\bot (k-1)}}{\|(s_{k})_{\bot (k-1)}\|}.\]
In our analysis, we set $\varphi_k$ to a zero-vector if $\|(s_{k})_{\bot (k-1)}\| < 10^{-4}$. 
Furthermore, we observe that in all our settings the evolved signaling systems generate constellation of signals that can be projected on a vector space of up to 2 dimensions. This allows us to represent the signals in a 2-D plane by simply computing the projection of each signal on $\varphi_{1}$ and $\varphi_{2}$.

\begin{figure}[ht!] 
  \centering
  \subfloat[Generation 1]{\label{fig:const_gen1}\includegraphics[width=0.24\textwidth]{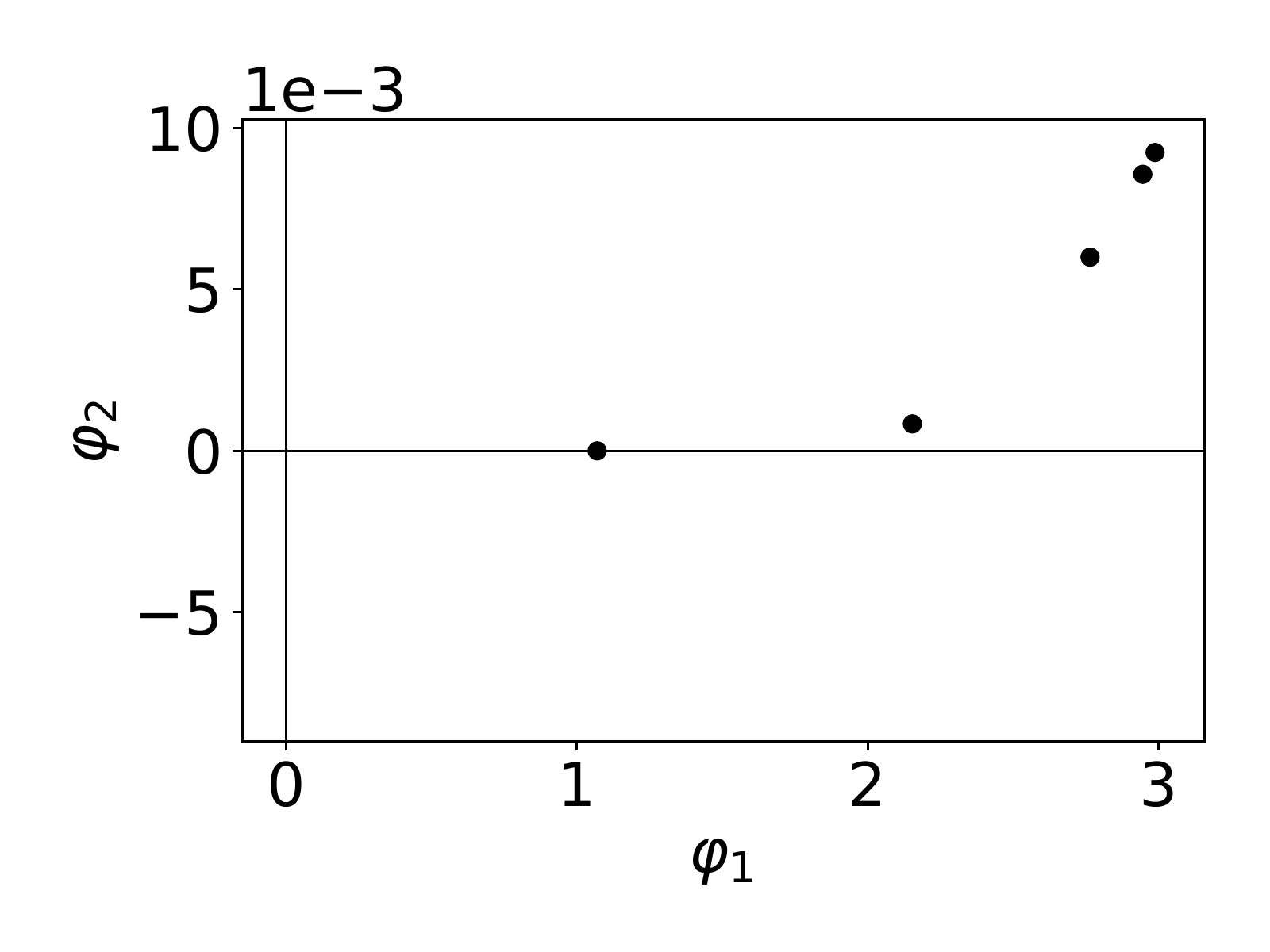}}
  \subfloat[Generation 2]{\label{fig:const_gen2}\includegraphics[width=0.24\textwidth]{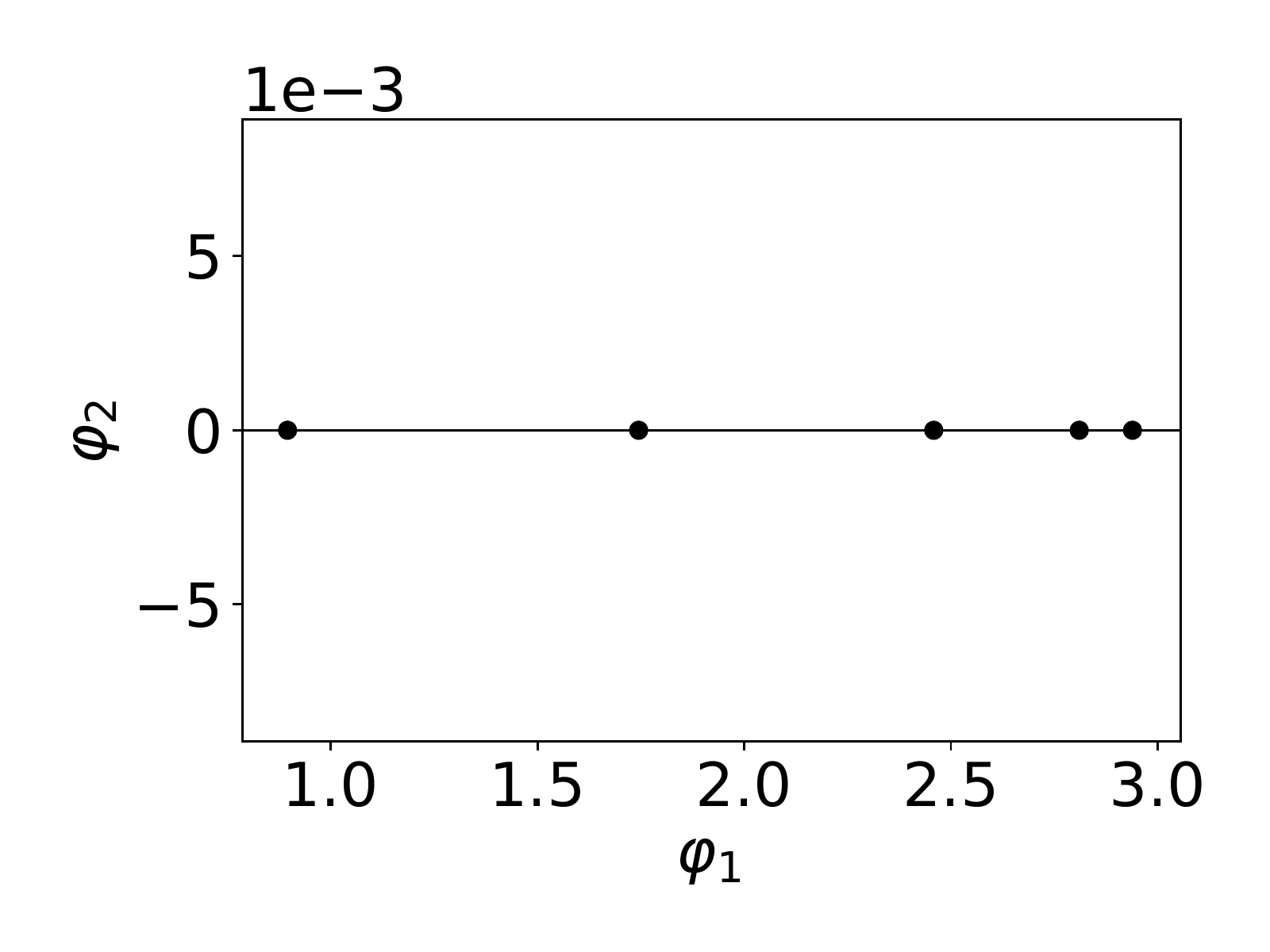}}

  \subfloat[Generation 5]{\label{fig:const_gen5}\includegraphics[width=0.24\textwidth]{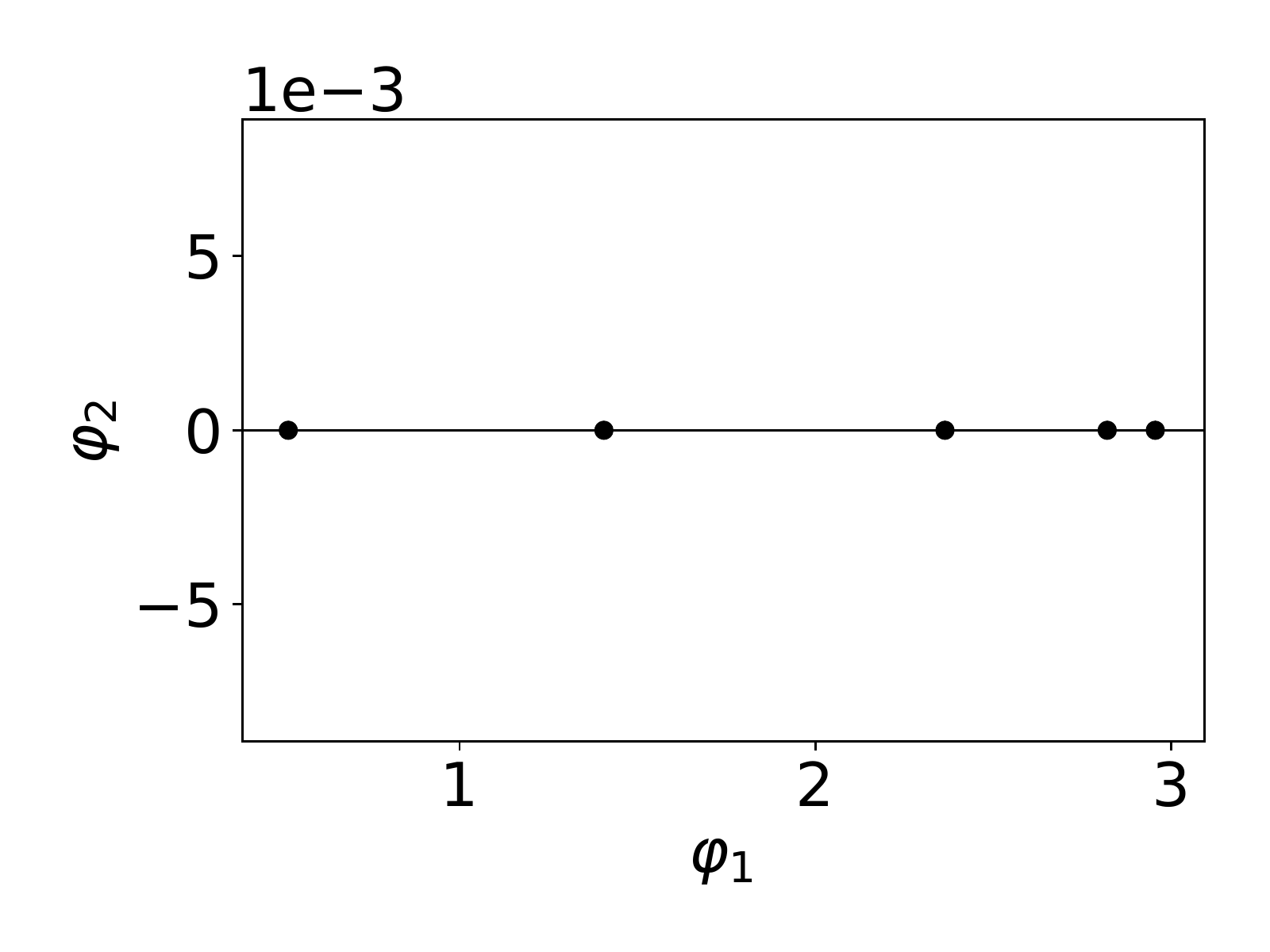}}
  \subfloat[Generation 6]{\label{fig:const_gen6}\includegraphics[width=0.24\textwidth]{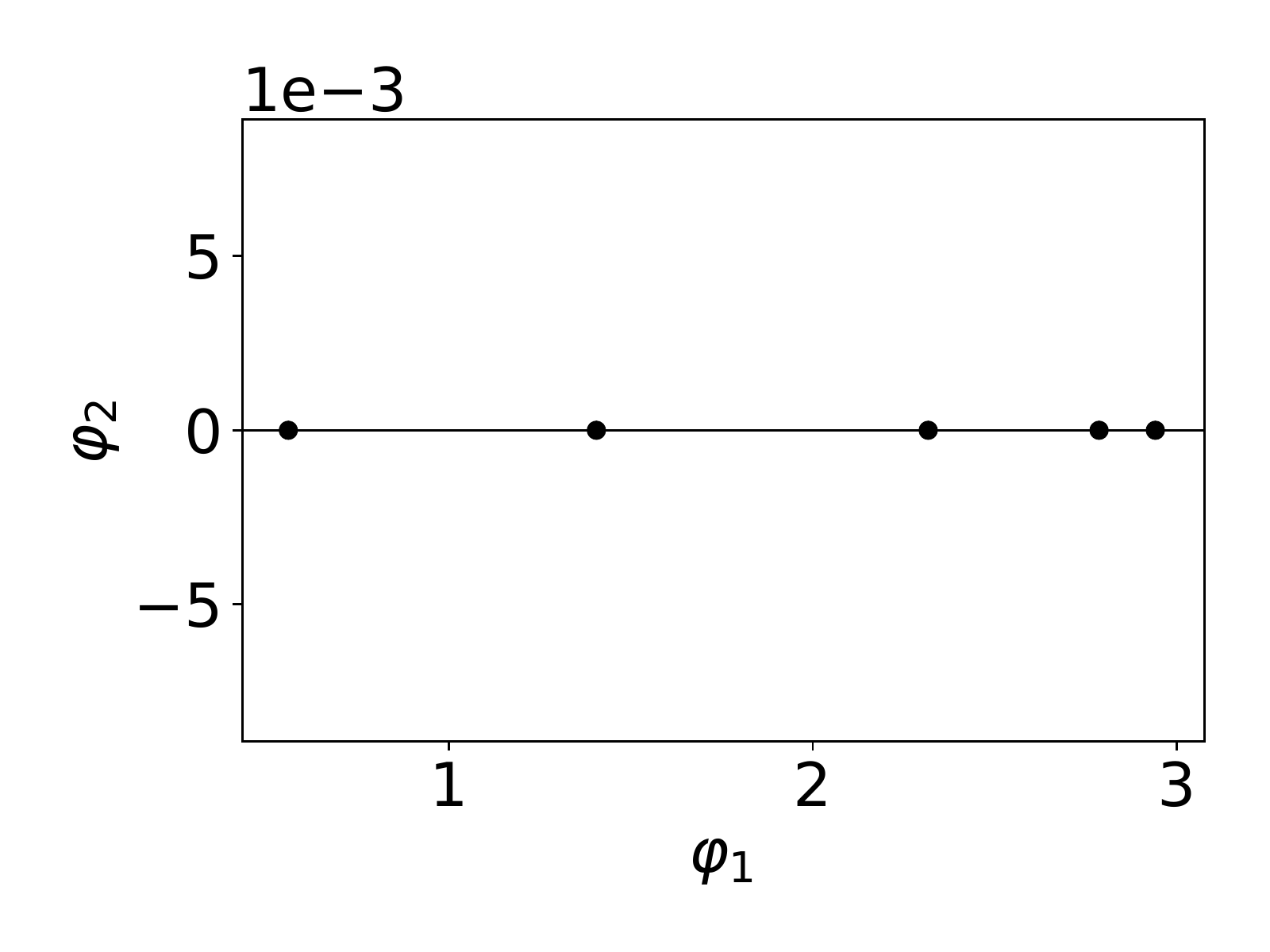}}
  
  \subfloat[Generation 11]{\label{fig:const_gen11}\includegraphics[width=0.24\textwidth]{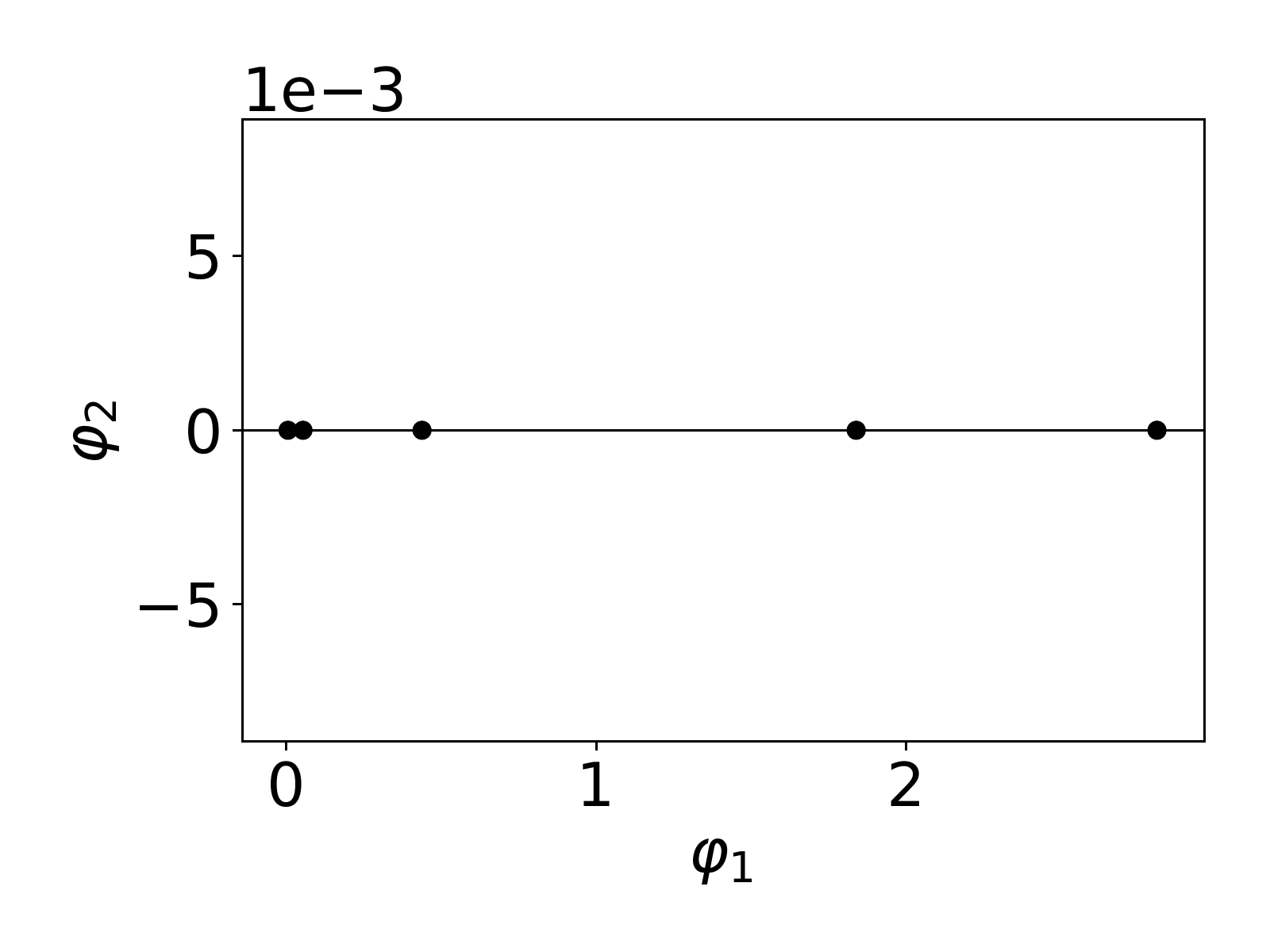}}
  \subfloat[Generation 45]{\label{fig:const_gen45}\includegraphics[width=0.24\textwidth]{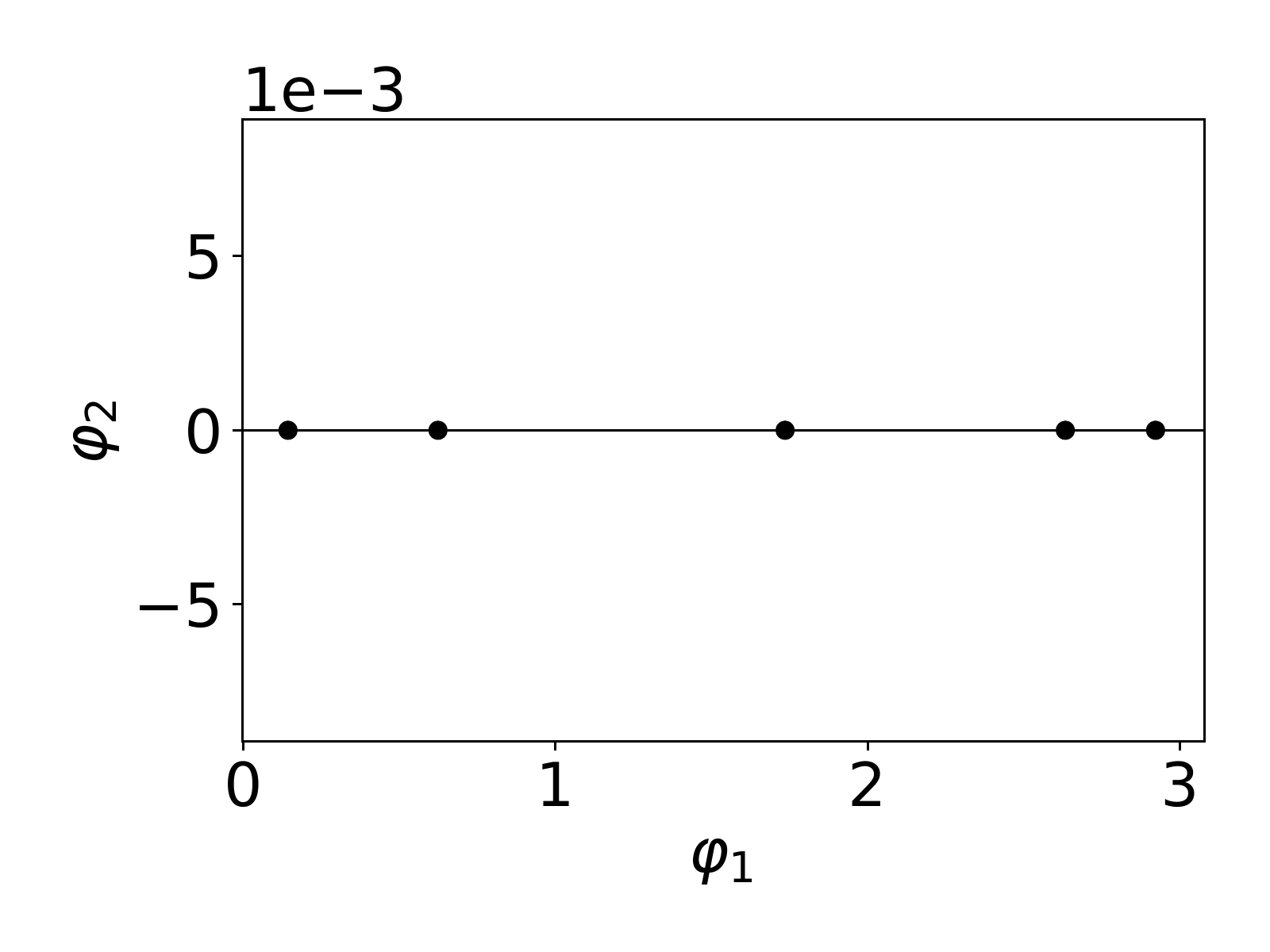}}
  \caption{\label{fig:constellations}Constellation of signals of the best signaling system evolved during a typical evolutionary run on the regression setting with limited amplitude, vocabulary of size 5, and without noise.}
\end{figure}

A typical evolutionary process (in the considered regression setting) proceeds as follows:
\begin{itemize}[leftmargin=*]
    \item In the first generation (\cref{fig:const_gen1}), the signals lie on a plane. As we can see, the margin between signals is small, meaning that some pairs of signals are less easy to distinguish from others.
    \item Then, for a certain number of generations (Figures \ref{fig:const_gen2}-\ref{fig:const_gen11}) the signals lie on a line (i.e., there are no orthogonal components, somewhat similar to a PAM system \cite{Barry2004}) but, as in the first generation, there are still pairs of signals separated by a small margin (with respect to other pairs).
    \item Eventually (\cref{fig:const_gen45}), the spacing between signals increases, leading to a more robust signaling system (optimal zero-fitness).
\end{itemize}

It is important to note that in this specific setting the best signaling system does not necessarily correspond to a system with equally spaced signals. This is due to the absence of noise, which entails that an equal margin between the signals does not constitute an evolutionary advantage. 

\begin{figure}[ht!]
    \centering
    \includegraphics[scale=0.5]{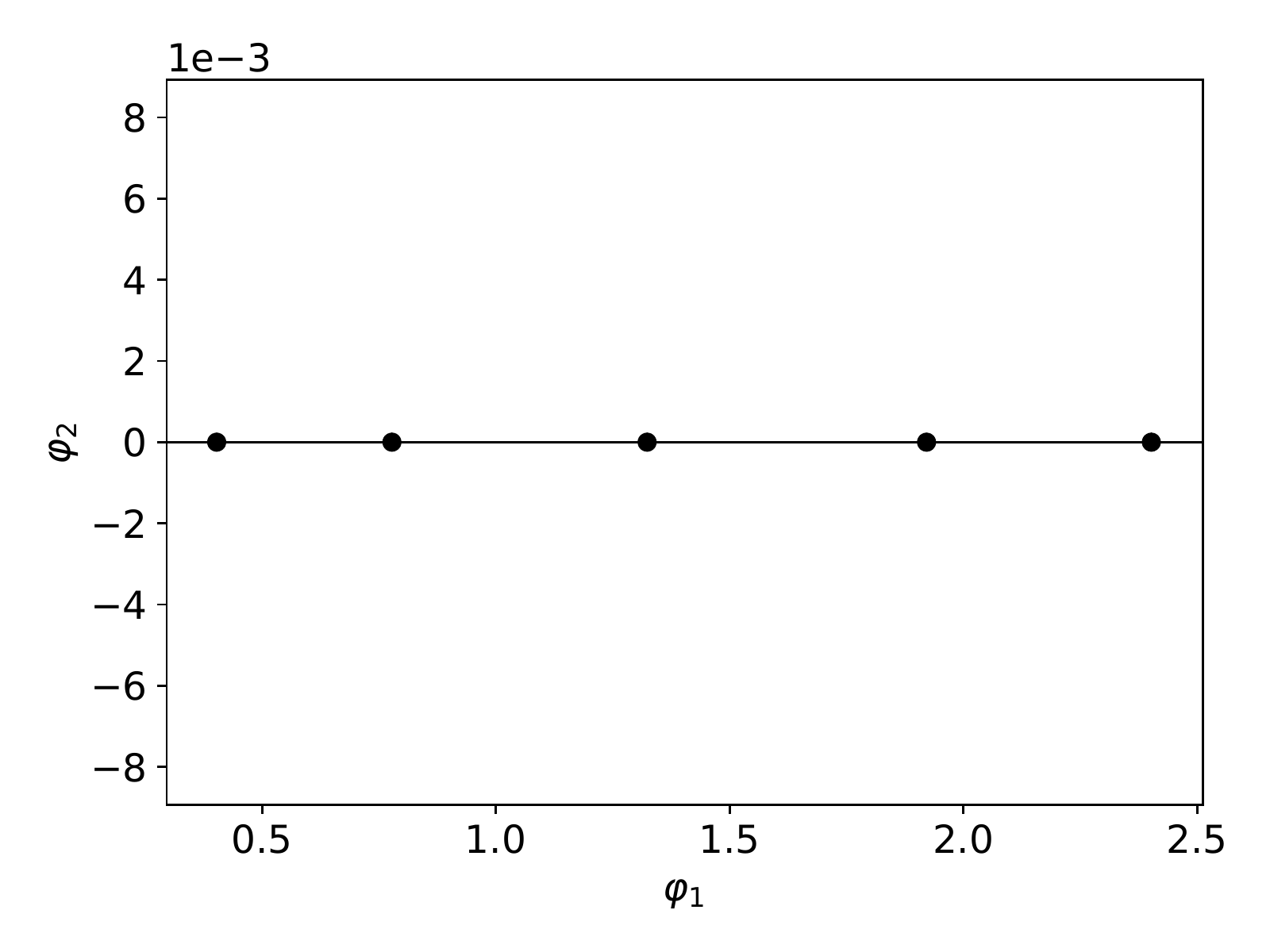}
    \caption{Constellation of signals of the best signaling system evolved on the regression setting with limited amplitude, vocabulary of size 5, and noise ($\sigma=0.1$, 3 trials).}
    \label{fig:constellation_noise}
\end{figure}

In comparison, the same analysis on one of the runs of the regression case with limited amplitude in the presence of noise ($\sigma=0.1$) with $3$ trials per concept (results shown in \cref{tab:noise3}) reveals that, while in the first part of the evolutionary process the constellations are similar to the ones shown in \cref{fig:const_gen1}-\ref{fig:const_gen11}, eventually the signaling system at the end of the evolutionary process (\cref{fig:constellation_noise}) consists in almost-equally spaced signals that allow a better adaptation to the noise present on the channel. As mentioned earlier, the analysis reported in the Supplementary Material for the other cases with noise show similar results.

\subsection{Clustering of the evolved signaling systems}
To understand whether the evolved signaling systems show common features, we perform a clustering of the best signaling systems evolved in the $20$ runs of the settings tested in \cref{sec:symbolic_comm}. We exclude from this analysis the classification with limited amplitude setting, due to the different vocabulary size w.r.t. the other settings. Since our goal is to discover if and how many clusters of signaling systems exist, we apply a clustering algorithm that, as opposed to non-density-based approaches (e.g. k-means), does not require to fix the number of clusters. Specifically, we use OPTICS \cite{ankerst1999optics}, a density-based clustering approach available in \texttt{scikit-learn} \cite{pedregosa2011scikit}. The parameters we used for the algorithm are: \texttt{min\_samples} = 4; \texttt{metric} = Chebyshev distance; \texttt{xi} = 0.1. All the other parameters are set to their default values.

\begin{figure}[ht!]
    \centering
    \includegraphics[scale=0.35,trim=0 0 0 0,clip]{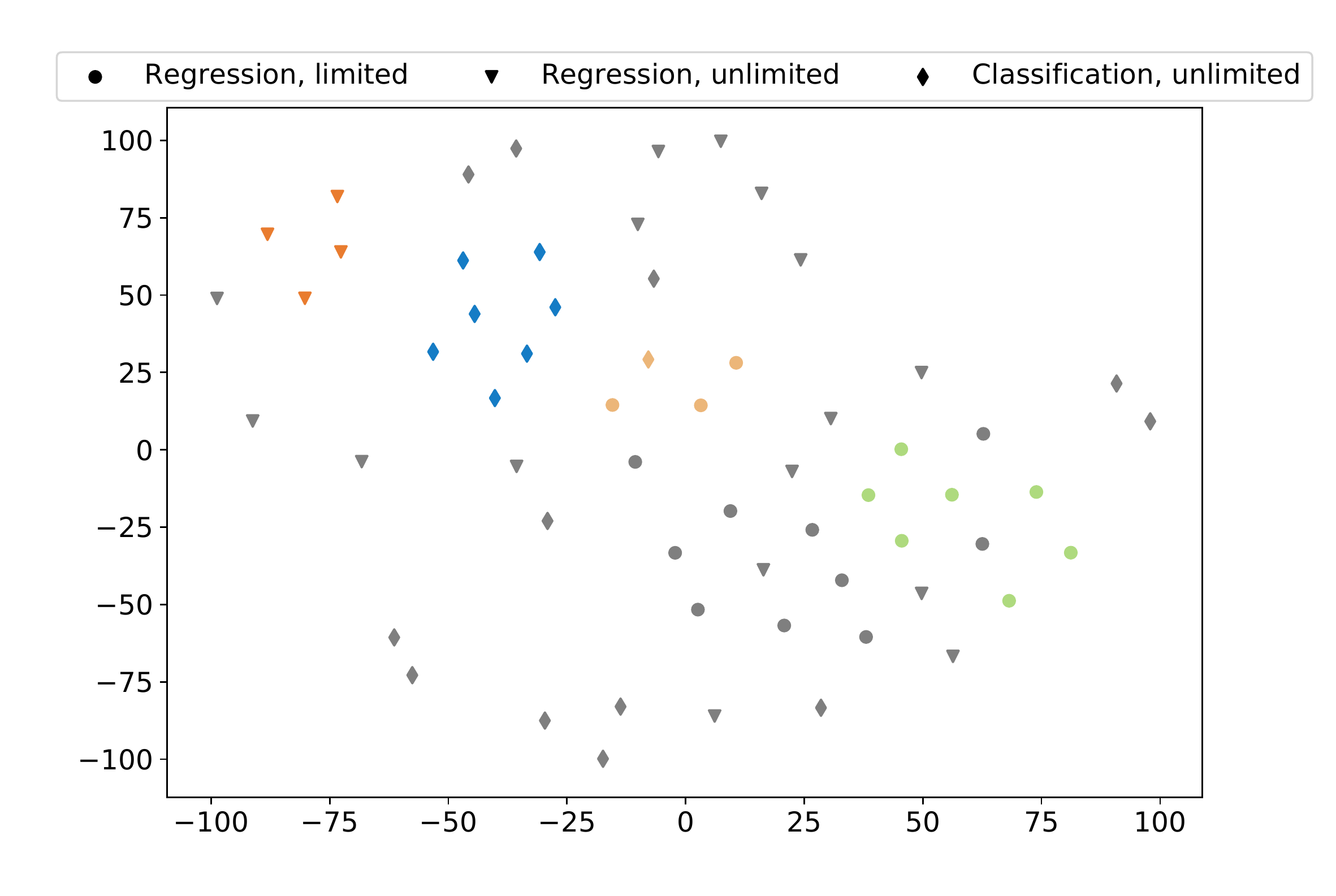}
    \caption{The evolved signaling systems in a 2-D vector space obtained with t-SNE. Each color represents a cluster (points classified as ``noise'' are in gray). The central signal for each cluster are represented in \cref{fig:central-signals}: green (\cref{fig:central_cluster_1}); yellow (\cref{fig:central_cluster_2}); blue (\cref{fig:central_cluster_3}); orange (\cref{fig:central_cluster_4}).
    }
    \label{fig:tsne-clustering}.
\end{figure}

\begin{figure}[ht!]
  \centering
  \subfloat[First cluster]{\label{fig:central_cluster_1}\includegraphics[width=0.24\textwidth]{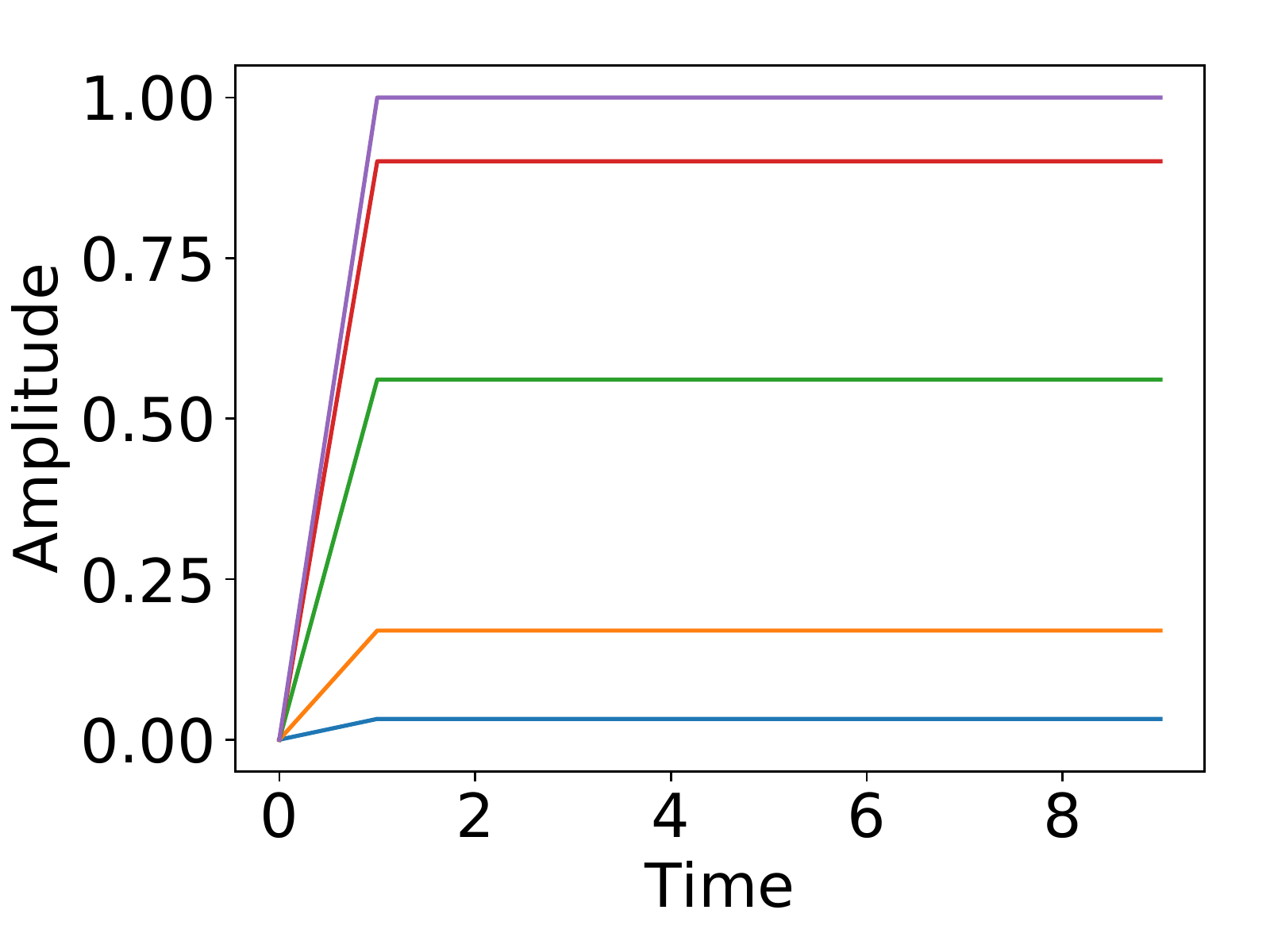}}
  \subfloat[Second cluster]{\label{fig:central_cluster_2}\includegraphics[width=0.24\textwidth]{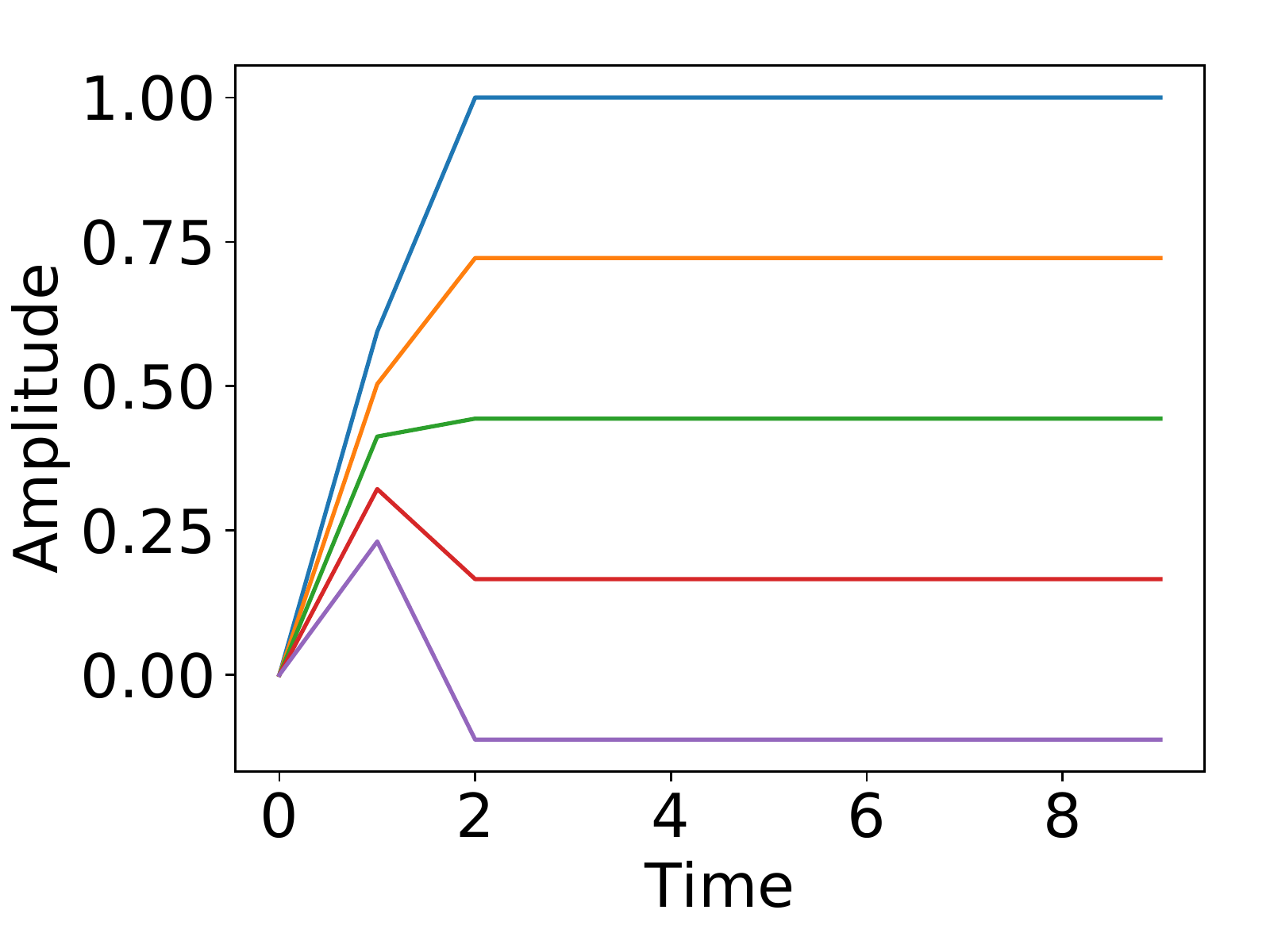}}

  \subfloat[Third cluster]{\label{fig:central_cluster_3}\includegraphics[width=0.24\textwidth]{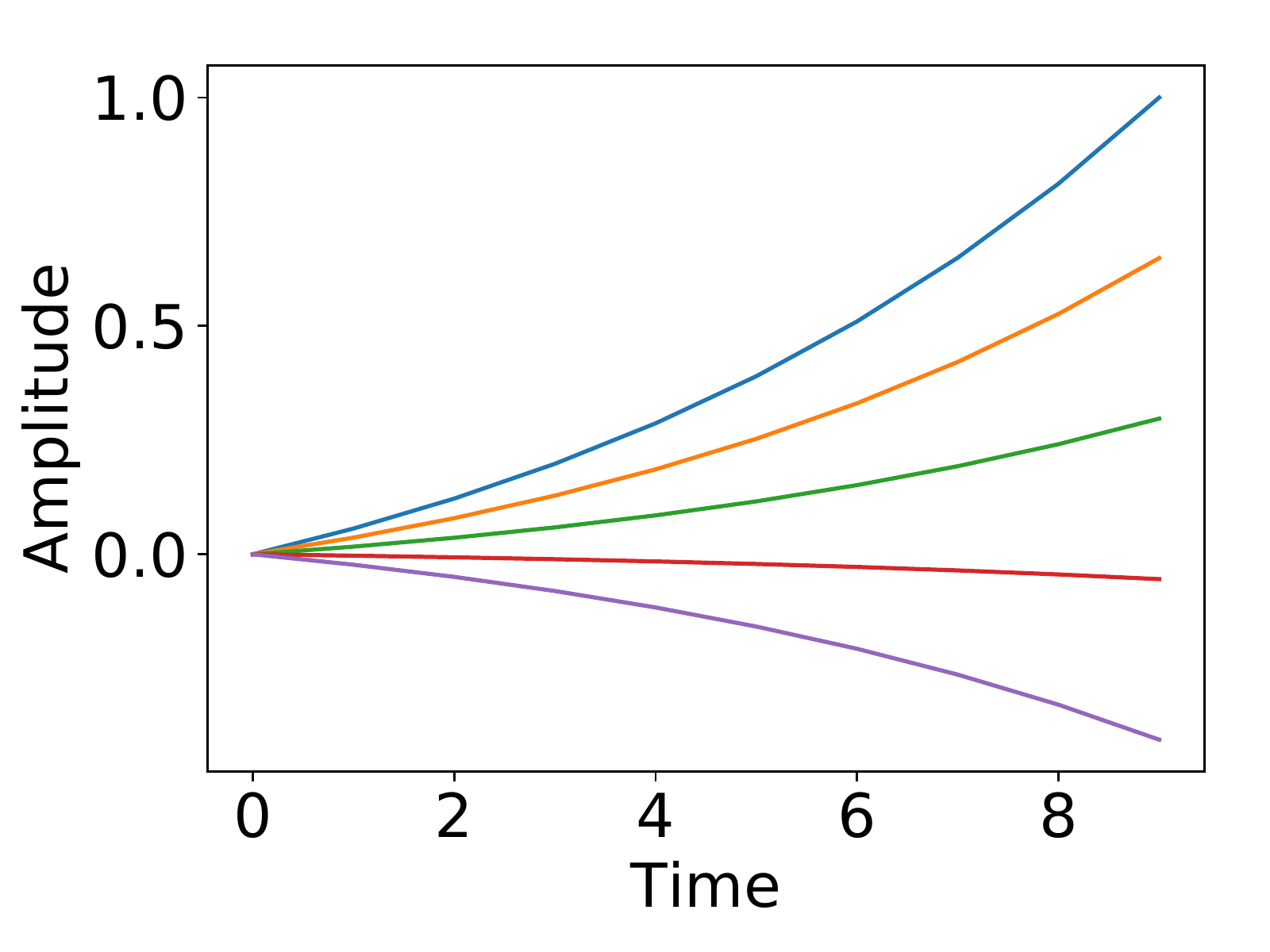}}
  \subfloat[Fourth cluster]{\label{fig:central_cluster_4}\includegraphics[width=0.24\textwidth]{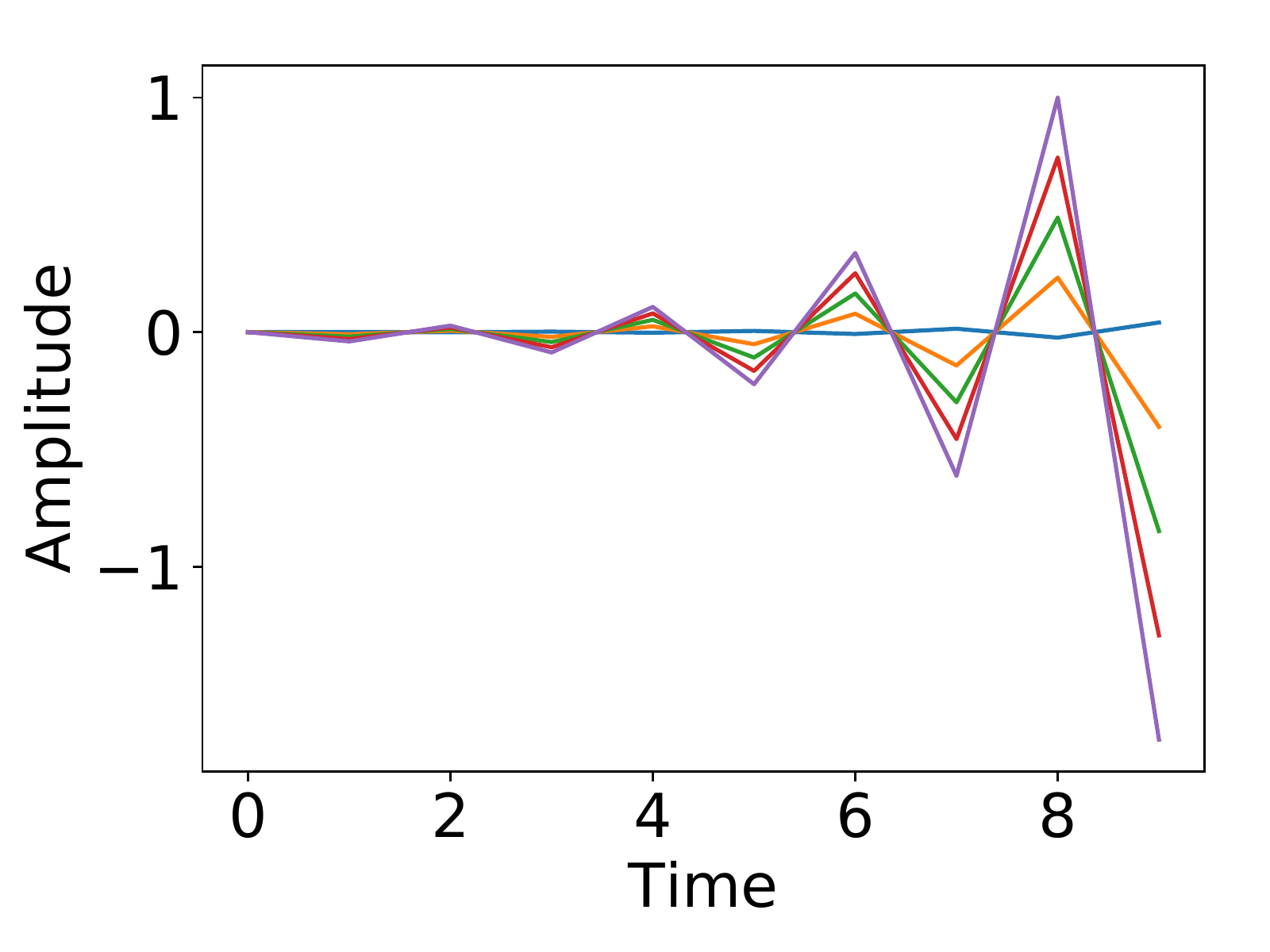}}
  \caption{\label{fig:central-signals}Central signals for each cluster shown in \cref{fig:tsne-clustering}.}
  \vspace{-1cm}
\end{figure}

The result of the clustering is shown in \cref{fig:tsne-clustering}, where we present a 2-D representation of the data by using the t-SNE algorithm \cite{maaten2008visualizing}.
We can see that most of the clusters contain signaling systems resulting from the same setting. Furthermore, for each cluster we compute the central signal by selecting the point that minimizes the mean distance from the other points in the cluster. The central signals are shown in \cref{fig:central-signals}. We observe four clusters of signals, of which three appear more clearly identifiable: the first one (\cref{fig:central_cluster_1}) contains signals that become constant at the first timestep; the third one (\cref{fig:central_cluster_3}) contains signals with an exponential-like shape; the fourth one (\cref{fig:central_cluster_4}) contains signals that look like triangular waves. On the other hand, the second cluster (\cref{fig:central_cluster_2}) is less clearly defined. In fact, it mostly contains signals that become constant after the second timestep, but also signals that look like the ones present in the first cluster.

Finally, by performing nearest neighbour classification (using the Chebyshev distance) of the signals we obtain that: $53.\overline{3}\%$ of the signaling systems are in the first cluster; $13.\overline{3}\%$ in the second one; $23.\overline{3}\%$ in the third one; $10\%$ in the forth one. Thus the first cluster appears to be a strong attractor for the evolutionary process.



\section{Conclusions}\label{sec:conclusions}
In this paper, we studied the evolution of symbolic communication in two contexts: a simple symbolic communication task and a a referential game. The agents communicate through waveforms sent on a dedicate channel. We analyzed the communication systems evolved in several settings under different points of view: convergence time, generalization capabilities, robustness to noise and properties of the evolved signals.

We found that the definition of the problem (i.e., classification or regression) has a significant impact on both the number of generations required to solve the task and the generalization capabilities. Furthermore, we observed that adding noise to the channel leads to communication systems that try to maximize the ``distance'' between the signals. Finally, we observed that the evolutionary process usually led to similar types of signaling systems, which can be grouped in four main categories, of which one appears dominant.

This study has, of course, some limitations. The first one concerns the agent controller. In fact, while the results obtained show that forms of symbolic communication can be evolved by using continuous-time recurrent neural networks, other forms of controllers/function approximators based e.g. on vanilla RNNs, Genetic Programming or Markov Brains may not necessarily lead to similar results. Another limitation regards the simplicity of our approach w.r.t. the evolution of communication in real-world scenarios. For instance, in our work we define and constrain one channel of communication, while in the real world communication may occur on multiple channels, and its effectiveness may depend on the context. Nevertheless, we believe that some of our findings (e.g. on the complexity of the problem and the effect of modelling it as either classification or regression) may hold true also in real-word scenarios and can be extended also to other controller architectures.  
Future work will address these two issues, and possibly focus on more complex (e.g. two-way) communication tasks as well as adversarial perturbations on the signals.


\balance

\bibliographystyle{ACM-Reference-Format}
\bibliography{main}

\end{document}